\documentclass[journal]{IEEEtran}
\usepackage{amsmath,amsfonts}
\usepackage{algorithmic}
\usepackage{algorithm}
\usepackage{array}
\usepackage[caption=false,font=normalsize,labelfont=sf,textfont=sf]{subfig}
\usepackage{textcomp}
\usepackage{stfloats}
\usepackage{url}
\usepackage{verbatim}
\usepackage{graphicx}
\usepackage{cite}

\usepackage{hyperref}
\usepackage{booktabs}       % professional-quality tables
\usepackage{nicefrac}       % compact symbols for 1/2, etc.
\usepackage{microtype}      % microtypography
\usepackage{graphicx}
\usepackage{amssymb}
\usepackage{multirow} 
\usepackage{colortbl}
\usepackage{pifont}
\newcommand{\cmark}{\ding{51}}%
\newcommand{\xmark}{\ding{55}}%
\usepackage[capitalize,noabbrev]{cleveref}
\usepackage{wrapfig}
\usepackage[dvipsnames]{xcolor}

\begin{document}

\title{Advancing Generalization in PINNs through Latent-Space Representations}

\author{Honghui Wang, Yifan Pu, Shiji Song,~\IEEEmembership{Senior Member,~IEEE}, Gao Huang,~\IEEEmembership{Member,~IEEE}
        % <-this % stops a space
\thanks{The work was supported by the National Natural Science Foundation of China under Grants 62276150. \textit{(Corresponding author: Gao Huang.)}}
\thanks{Honghui Wang, Yifan Pu, Shiji Song, and Gao Huang are with the
Department of Automation, Tsinghua University, Beijing 100084, China. \quad Email: \{wanghh20, puyf23\}@mails.tsinghua.edu.cn, shijis@mail.tsinghua.edu.cn, gaohuang@tsinghua.edu.cn}% <-this % stops a space
\thanks{This work has been submitted to the IEEE for possible publication.
Copyright may be transferred without notice, after which this version may
no longer be accessible.}
}

% The paper headers
% \markboth{Journal of \LaTeX\ Class Files,~Vol.~14, No.~8, August~2021}%
% {Shell \MakeLowercase{\textit{et al.}}: A Sample Article Using IEEEtran.cls for IEEE Journals}

% \IEEEpubid{0000--0000/00\$00.00~\copyright~2021 IEEE}
% Remember, if you use this you must call \IEEEpubidadjcol in the second
% column for its text to clear the IEEEpubid mark.

\maketitle
% \IEEEpeerreviewmaketitle
\newcommand{\name}{P\textsc{i}D\textsc{o}}
\begin{abstract}
Physics-informed neural networks (PINNs) have made significant strides in modeling dynamical systems governed by partial differential equations (PDEs). However, their generalization capabilities across varying scenarios remain limited.
To overcome this limitation, we propose {\name}, a novel physics-informed neural PDE solver designed to generalize effectively across diverse PDE configurations, including varying initial conditions, PDE coefficients, and training time horizons. {\name} exploits the shared underlying structure of dynamical systems with different properties by projecting PDE solutions into a latent space using auto-decoding. It then learns the dynamics of these latent representations, conditioned on the PDE coefficients.
Despite its promise, integrating latent dynamics models within a physics-informed framework poses challenges due to the optimization difficulties associated with physics-informed losses. To address these challenges, we introduce a novel approach that diagnoses and mitigates these issues within the latent space. This strategy employs straightforward yet effective regularization techniques, enhancing both the temporal extrapolation performance and the training stability of {\name}. 
We validate {\name} on a range of benchmarks, including 1D combined equations and 2D Navier-Stokes equations. Additionally, we demonstrate the transferability of its learned representations to downstream applications such as long-term integration and inverse problems.
\end{abstract}


\section{Introduction}
\label{sec:introduction}

Partial differential equations (PDEs) constitute the cornerstone of comprehending complex systems and forecasting their behavior.
Recent years have witnessed a surge in the effectiveness of deep learning methods for solving PDEs~\cite{yu2018deep,kovachki2021neural,brandstetter2021message}. 
Among these, Physics-Informed Neural Networks (PINNs) have emerged as a burgeoning paradigm~\cite{raissi2019physics}.
PINNs leverage Implicit Neural Representations (INRs) to parameterize PDE solutions, enabling them to effectively bridge data with mathematical models and tackle high-dimensional problems. 
This unique characteristic has led to their widespread adoption in a variety of applications, 
including computational fluid dynamics~\cite{raissi2020hidden}, 
smart healthcare~\cite{oszkinat2022uncertainty} and intelligent manufacturing~\cite{hua2023physics,wang2023inherently}.

A key advantage of PINNs lies in their ability to be trained by enforcing PDE-based constraints even in the absence of exact solutions. 
This flexibility proves valuable in real-world settings where perfect data might not be available. 
However, this very benefit comes at a cost. Each instance of PINNs is trained tailored to a specific configuration of initial and boundary conditions, PDE coefficients, geometries, and forcing terms.
Modifying any of these elements necessitates retraining, resulting in significant computational inefficiency. 
In addressing this obstacle,
Neural Operators (NOs) have been proposed as a promising solution~\cite{li2020fourier,lu2021learning}. 
NOs aim to tackle the so-called \textit{parametric} PDEs by learning to map variable condition entities to corresponding PDE solutions. 

Despite their success, several limitations hinder the generalization ability of NOs. 
First, some NO architectures restrict the choice of grids. This limitation leads to challenges when encountering input ~\cite{lu2021learning,wang2021learning} or output~\cite{li2020fourier} grids that differ from those used in training. 
Secondly, for time-dependent PDEs, NOs are typically trained to provide predictions within a fixed time horizon, limiting their applicability in scenarios requiring broader temporal coverage. 
Finally, while NOs may generalize well to specific types of variable conditions, their ability to handle concurrent variations in multiple types of conditions has not been thoroughly validated. 
This is particularly concerning in domains like coefficient-aware dynamics modeling, where models must simultaneously adapt to varying initial conditions and PDE coefficients~\cite{brandstetter2021message}.

\begin{figure*}[t]
% \vskip 0.2in
\begin{center}
\centerline{\includegraphics[width=\linewidth]{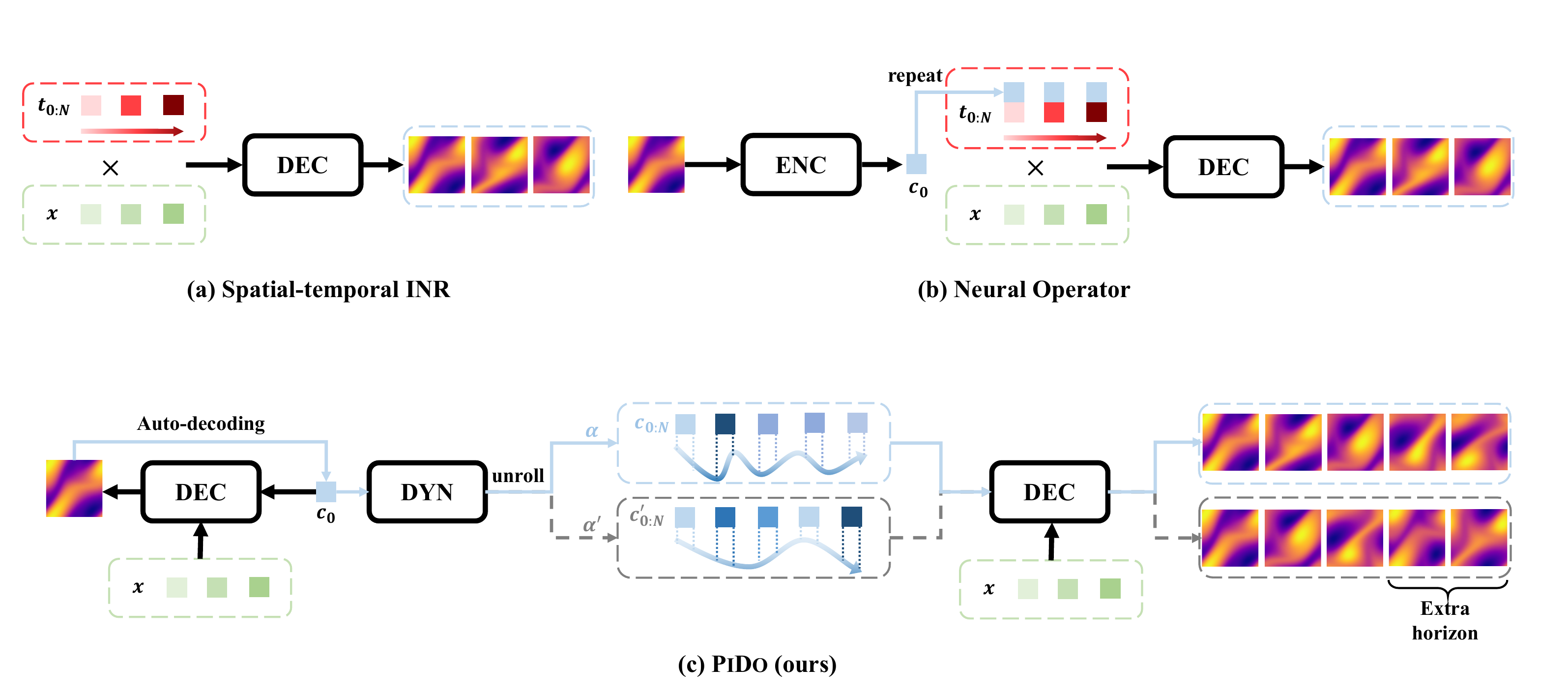}}
% \vskip -0.1in
\caption{Physics-informed neural PDE solvers. Colorful squares denote spatial/temporal coordinates (``$x$'' and ``$t$'') and embeddings of PDE solutions (``$c$'') at different times, while curves represent continuous trajectories of embeddings unrolled by the dynamics model, with colors indicating different values. The "$\times$" denotes Cartesian product. 
``ENC'', ``DEC'' and ``DYN'' stand for the encoder, decoder and dynamics model, respectively.
{\name} encodes PDE solutions into a compact latent space, enabling generalization to \textbf{unseen initial conditions}. It models and unrolls latent trajectories under \textbf{varying PDE coefficients} ($\boldsymbol{\alpha}$ and $\boldsymbol{\alpha'}$) using a dynamics model, leveraging its strength in \textbf{temporal extrapolation}. The entire framework is trained end-to-end with a physics-informed loss.
}
\label{fig:framework}
\end{center}
% \vskip -0.35in
\end{figure*}

This paper tackles the limited generalization ability in physics-informed neural PDE solvers.
To this end, we introduce the \textbf{P}hysics-\textbf{I}nformed \textbf{D}ynamics representati\textbf{O}n learner ({\name}), a versatile framework capable of generalizing across different types of variables in PDE configurations, including initial conditions, PDE coefficients and training time horizons.
The key to {\name}'s success lies in its two core components as shown in \cref{fig:framework}. 
First, the grid-independent \textit{spatial representation learner} (denoted as DEC) exploits the intrinsic structure shared between dynamical systems governed by different PDE coefficients. 
This is achieved by learning a mapping between the solution space and a low-dimensional latent space via auto-decoding~\cite{park2019deepsdf}. 
This latent space captures the essential representations of the solutions across all time steps, enhancing generalization to unseen initial conditions while also enabling more efficient dynamics modeling compared to operating in the high-dimensional data space.
Inspired by Explicit Dynamics Modeling (EDM)~\cite{yin2022continuous,wan2022evolve}, the \textit{temporal dynamics model} (denoted as DYN) then learns the coefficient-aware evolution of latent embeddings using Neural ODEs~\cite{chen2018neural}, which excel at generalizing across PDE coefficients and extrapolating beyond the training horizon.

While previous EDM methods are typically trained in a data-driven manner, their performance relies heavily on dataset size. In contrast, we focus on a physics-informed setting, where {\name} is optimized to satisfy governing PDEs without relying on real data. However, the physics-informed loss is known to face optimization challenges~\cite{krishnapriyan2021characterizing, wang2021understanding}, leading to two key issues in its integration with EDM: instability during training and degradation in time extrapolation. To address these challenges, we adopt a novel perspective by diagnosing and tackling them within the latent space. By projecting data into low-dimensional representations, we identify two problematic latent behaviors: overly complex dynamics and latent embedding drift. 
Based on these insights, we propose two regularization techniques-\textit{Latent Dynamics Smoothing} and \textit{Latent Dynamics Alignment}-to improve training stability and extrapolation ability, respectively. Overall, these strategies enhance {\name}’s generalization ability compared to its data-driven counterparts.
Our contributions can be summarized as follows:
\begin{itemize}
    \item We propose a novel physics-informed learning framework that achieves robust generalization across diverse variables in PDE configurations.
    \item We diagnose the learning difficulties of physics-informed dynamics models within the latent space and address them using latent dynamics smoothing and latent dynamics alignment, resulting in improved generalization performance compared to the data-driven counterpart.
    \item We demonstrate the effectiveness of {\name} on extensive benchmarks including 1D combined 
    equation and 2D Navier-Stokes equations, and explore the transferability of {\name}'s learned 
    representations to downstream tasks including long-term integration and inverse problems.
\end{itemize}

\section{Related Work}
\label{sec:related_work}

\begin{table}[t]
\caption{Comparisons of Physics-informed Neural PDE Solvers.}
\label{tab:comparison_methods}
% \vskip -0.5in
% \begin{center}
% \begin{small}
% \begin{sc}
\resizebox{1.0\linewidth}{!}{
\begin{tabular}{lcccc}
\toprule
 \multirow{2}{*}{Method}&  Initial conditions & PDE coefficients & Time & \multirow{2}{*}{Data-free}  \\
 & generalization & generalization & extrapolation&\\
\midrule
PINN\cite{raissi2019physics} & \color{red}\xmark & \color{red}\xmark & \color{red}\xmark & 
\color{teal}\cmark  \\
PI-DeepONet\cite{wang2021learning} &\color{teal}\cmark & \color{teal}\cmark & \color{red}\xmark & 
\color{teal}\cmark \\
MAD\cite{huang2022meta}  & \color{teal}\cmark & \color{teal}\cmark & 
\color{red}\xmark & 
\color{teal}\cmark  \\ 
PINODE\cite{sholokhov2023physics}  & \color{teal}\cmark & \color{red}\xmark & \color{teal}\cmark & \color{red}\xmark\\ 
DIN\sc{o}\cite{yin2022continuous}  & \color{teal}\cmark & \color{red}\xmark & \color{teal}\cmark & \color{red}\xmark \\ 
\name~(Ours) & \color{teal}\cmark & \color{teal}\cmark &  \color{teal}\cmark & \color{teal}\cmark \\
\bottomrule
\end{tabular}}
% \end{sc}
% \end{small}
% \end{center}
% \vskip -0.2in
\end{table}
\paragraph{Spatial-temporal Implicit Neural Representations} are powerful paradigm to model continuous signals in 3D shape representation learning and PDEs solving~\cite{sitzmann2020implicit,fathony2020multiplicative}. They train a neural network to map spatial-temporal coordinates to PDE solutions as
\begin{equation}
\label{eq:inr}
    \textstyle \mathcal{G}: (t,\boldsymbol{x}) \rightarrow \boldsymbol{u}(t,\boldsymbol{x}),
\end{equation}
enabling the model to query any point without being constrained by the resolution of a fixed grid.
Taking advantage of this differentiable parameterization, physics-informed neural networks (PINNs) embed PDEs as soft constraints to guide the learning process~\cite{yu2018deep,raissi2019physics,fang2021high,hillebrecht2023rigorous,kapoor2023physics,wu2024parallel}. This framework is appealing in many realistic situations as it can be trained in the absence of exact PDE solutions. However, it is well known that the PDE-based constraints suffer from ill-conditioned optimization~\cite{wang2021understanding}. Many recent works have devoted efforts to alleviate the optimization difficulty and to improve training efficiency with loss weight balancing~\cite{wang2022and,wang2022respecting,yao2023multi}, curriculum learning~\cite{krishnapriyan2021characterizing} and dimension decomposition~\cite{cho2023separable}. 
Another limitation of PINNs lies in its generalization ability, as the model can be only applied to a predefined set of PDE coefficients and initial conditions.

\paragraph{Neural Operators} attempt to learn a mapping between two infinite-dimensional function space as 
\begin{equation}
    \textstyle \mathcal{G}: (\mathcal{A}\times\mathbb{R}) \rightarrow \mathcal{U},\quad (\boldsymbol{a}, t) \rightarrow \mathcal{G}(\boldsymbol{a}, t)=\boldsymbol{u}(t).
\end{equation}
In the context of parametric PDEs, $\boldsymbol{a}\in\mathcal{A}$ is a function characterizing initial conditions or PDE coefficients and $\boldsymbol{u}\in\mathcal{U}$ denotes the corresponding solution. As per the inherent design, NOs are grid-independent~\cite{li2020multipole,li2020fourier,lu2021learning}. 
However, many NOs do not have full flexibility on the spatial discretization.
For example, DeepONets~\cite{lu2021learning} use an INR to represent the continuous solution, 
% FNO is restrict to uniform Cartesian grids and 
but reply on a fixed discretization of the input function $\boldsymbol{a}$. Additionally, they require a large number of parameter-solution pairs to learn the solution operator. To address the need of solution data, PI-DeepONet~\cite{wang2021learning} proposes to incorporate neural operators with physics-informed training. It can be seen as an extension to PINNs by conditioning the output linearly on the embeddings of input parameters encoded by a branch network. However, this linear aggregation strategy limits its capability in handling complex problems~\cite{lanthaler2022nonlinear}. As an alternative approach, Meta-Auto-Decoder (MAD) learns an embedding for each input parameter via auto-decoding~\cite{huang2022meta}. 
Another line of work~\cite{li2021physics} approximates the PDE-based loss with finite difference method based on the outputs of NOs, which is prone to the discretization error. 
Finally, a key limitation of NOs for time-dependent PDEs is their inherent design for prediction within a specific horizon, $[0,T_{tr}]$, which restricts generalization beyond the time $T_{tr}$.

\paragraph{Explicit Dynamics Modeling} learns the derivative of solutions with respective to time instead of directly fitting solution values at different time steps as spatial-temporal INRs and NOs do. The solution at a given time can be obtained with numerical integration, which can be formalized as 
\begin{equation}
\label{eq:edm}
    \textstyle \mathcal{G}: \boldsymbol{u} \rightarrow \frac{\partial \boldsymbol{u}}{\partial t},\quad \boldsymbol{u}(t)=\boldsymbol{u}(0)+\int_{\tau=0}^t\mathcal{G}(\boldsymbol{u}(\tau))d\tau.
\end{equation}
One kind of EDM is autoregressive models~\cite{greenfeld2019learning,brandstetter2021message}. They learn to predict the solution at $t+\delta t$ based on the current solution at $t$ with $\mathcal{G}: \boldsymbol{u}(t) \rightarrow \boldsymbol{u}(t+\delta t)$, 
which equals to approximating the time derivative with finite difference$\frac{\boldsymbol{u(t+\delta t)}-\boldsymbol{u(t)}}{\delta t}$. 
Another kind of EDM learns the time derivative with Neural ODEs~\cite{chen2018neural}. Neural ODEs can be queried at any time step and have been widely applied in continuous-time modeling~\cite{quaglino2019snode,ayed2019learning}.  
EDM has shown superior extrapolation ability beyond training time interval compared with INRs and NOs in solving initial value problems~\cite{yin2022continuous,wan2022evolve,serrano2023operator}. In spite of its efficacy, EDM has two main limitations. First, most EDM methods are agnostic to the underlying PDEs, restricting them to modeling a single type of dynamics. Second, the training of EDM can be data-intensive, especially when modeling multiple dynamics. In this case, the required dataset size grows exponentially with the number of PDE coefficients. 

Our proposed method builds upon EDM and addresses these challenges by incorporating physics-informed training. This approach leverages the strengths of both methods 
(cf.~\cref{tab:comparison_methods}). 
Furthermore, we show that the latent space introduced by EDM offers a novel perspective for diagnosing learning difficulties in physics-informed loss.
While prior work has explored combining physics-informed training with EDM, these efforts have limitations. PINODE~\cite{sholokhov2023physics}, for example, utilizes an auto-encoder structure confined to a fixed grid. 
Additionally, its physics-informed training requires the input data to be sampled from a pre-defined analytical distribution that is similar to the true PDE solutions distribution (see~\cref{sec:detail_auto_encoder} for details). This requirement is often impractical in real-world scenarios. In contrast, our method does not require any prior assumptions about the data distribution. Another work~\cite{wen2023reduced} cannot be trained solely with the physics-informed loss.

\section{Method}
\label{sec:method}
In this section, we first outline the problem setting in \cref{sec:method_setup}, followed by the architectural design of {\name} in \cref{sec:model}. Next, we introduce the physics-informed training adapted for EDM in \cref{sec:optimization} and analyze the related learning challenges within the latent space in \cref{sec:diagnose}.
We provide pseudo-code for training and testing procedures in~\cref{alg:train} and~\cref{alg:test}, respectively.

\subsection{Problem Setup}
\label{sec:method_setup}
We focus on the time-dependent parametric PDEs which can be formulated as 
\begin{equation}
\label{eq:parametirc_pde}
\begin{split}
    &\frac{\partial\boldsymbol{u}(t,\boldsymbol{x})}{\partial t}+\mathcal{L}^{\boldsymbol{\alpha}}(\boldsymbol{u}(t,\boldsymbol{x}))=0,\ \ (t,\boldsymbol{x})\in[0,T]\times\Omega,\\
    &\mathcal{B}(\boldsymbol{u}(t,\boldsymbol{x}))=0,\ \ \ \ \ \ \ \ \ \ \ \ \ \ \ \ \ \ (t,\boldsymbol{x})\in[0,T]\times\partial\Omega, \\
    &\boldsymbol{u}(t=0,\boldsymbol{x})=\boldsymbol{\phi}(\boldsymbol{x}),\ \ \ \ \ \ \ \ \ \ \ \ \ \boldsymbol{x} \in \Omega,
\end{split}
\end{equation}
where $\mathcal{L}^{\boldsymbol{\alpha}}$ is the differential operator parameterized by coefficients $\boldsymbol{\alpha}$ and $\boldsymbol{u}:[0,T]\times\Omega \rightarrow \mathbb{R}^n$ is the solution. Let $\boldsymbol{u}_t \triangleq \boldsymbol{u(t)}$ denotes the state of interest at each time step $t$. 
Our goal is to solve the parametric PDEs by learning the map $\mathcal{G}^*: (t,\boldsymbol{\phi},\boldsymbol{\alpha}) \rightarrow \boldsymbol{u}_t$ from parameters to solutions. The model is trained on a set of parameters $\{(\boldsymbol{\phi}_i,\boldsymbol{\alpha}_i)\}$ sampled from distribution $\Phi$. At test time, we evaluate the model on unseen initial conditions and coefficients sampled from the same distribution. In addition, we assess the model's temporal extrapolation capability. To achieve this, the training horizon is confined to the interval $[0, T_{tr}]$, where $T_{tr} < T$, and the corresponding test horizon, denoted as $[0, T_{ts}]$, spans the entire time interval with $T_{ts} = T$.

\subsection{Model Architecture}
\label{sec:model}

The proposed {\name} tackles the dynamics modeling task defined in~\cref{eq:parametirc_pde} by approximating the solution map $\mathcal{G}^*: (t,\boldsymbol{\phi},\boldsymbol{\alpha}) \rightarrow \boldsymbol{u}_t$ with  
\begin{equation}
% \vskip -0.1in
\label{eq:unroll}
\textstyle \boldsymbol{\tilde{u}}_t(\boldsymbol{x}) = \mathcal{D}\big(\boldsymbol{c}_t, \boldsymbol{x}\big),\, \text{where}\, \boldsymbol{c}_t=\mathcal{E}(\boldsymbol{\phi}) + \int_{\tau=0}^t\mathcal{F}(\boldsymbol{c}_{\tau}, \boldsymbol{\alpha}) d\tau. 
\end{equation}
{\name} achieves this with two key components as shown in~\cref{fig:framework}. First, the spatial representation learner establishes a mapping between the data space and a low-dimensional representation space. It consists of an encoder $\mathcal{E}$, which transforms initial conditions $\boldsymbol{\phi}$ into the latent space via auto-decoding~\cite{park2019deepsdf}, and a decoder $\mathcal{D}$, which represents continuous solutions given a latent embedding $\boldsymbol{c}_t$ at time step $t$.  
Second, the temporal dynamics model $\mathcal{F}$ learns the coefficients-aware evolution of latent embeddings starting from the initial embedding $\mathcal{E}(\boldsymbol{\phi})$. 

\paragraph{Spatial representation learner}
aims to capture the essential representations of PDE solutions in a low-dimensional latent space. 
It achieves this by learning a decoder $\mathcal{D}$ which can accurately reconstruct solution trajectories $\boldsymbol{u}$ from low-dimensional embeddings $\boldsymbol{c}$. 
We parameterize the decoder $\mathcal{D}$ with an INR, which approximates the solution $\boldsymbol{u}$ conditioned on both spatial coordinates and embeddings, resulting in $ \boldsymbol{\tilde{u}}(\boldsymbol{x})=\mathcal{D}(\boldsymbol{c}, \boldsymbol{x})$. Benefiting from this parameterization, {\name} achieves grid-independence (see~\cref{sec:related_work}) and can readily compute the spatial derivatives of $\boldsymbol{\tilde{u}}$ through Automatic differentiation (AD)~\cite{baydin2018automatic}, which are crucial for physics-informed training.

Given a learned decoder $\mathcal{D}$, the encoder $\mathcal{E}$ is defined via auto-decoding. Specifically, the encoder $\mathcal{E}$ identifies the corresponding embedding $\boldsymbol{c}$ of spatial observation $\boldsymbol{u}$ (or initial condition $\boldsymbol{\phi}$) through an optimization process.
This process minimizes the reconstruction error between $\boldsymbol{u}$ and $\boldsymbol{\tilde{u}}=\mathcal{D}(\boldsymbol{c})$ by updating a learnable $\boldsymbol{c}$.
In essence, the auto-decoding seeks the optimal embedding $\boldsymbol{c}^*$ that captures the essential information within $\boldsymbol{u}$ necessary for the decoder to accurately reproduce the original observation. This encoder $\mathcal{E}$ can be formulated as 
\begin{equation}
\label{eq:auto-decoder}
\boldsymbol{c}^*=\mathcal{E}(\boldsymbol{u})  , \text{where}\  \boldsymbol{c}^* = \operatorname*{argmin}_{\boldsymbol{c}} \mathbb{E}_{\boldsymbol{x}\in\Omega}\|\boldsymbol{u}(\boldsymbol{x}) -\mathcal{D}(\boldsymbol{c}, \boldsymbol{x})\|^2.
\end{equation}  
Here the expectation (denoted by $\mathbb{E}$) is taken over spatial coordinates $\boldsymbol{x}$ sampled the domain $\Omega$.
In practice, this optimization process can be efficiently achieved by updating the latent embedding $\boldsymbol{c}$ (typically initialized with zeros in our experiments) with a few steps of gradient descent.

Our method distinguishes itself from prior works~\cite{wang2021learning,huang2022meta} by learning the latent space of entire trajectories, 
not just initial conditions. 
This enables us to thoroughly capture the intrinsic structure of solution space, leading to enhanced generalization across initial conditions. 

\paragraph{Temporal dynamics model}  leverages a Neural ODE to learn the evolution of latent embeddings as
\begin{equation}
\label{eq:node}
\textstyle \frac{\partial\boldsymbol{c}_t}{\partial t} =\mathcal{F}(\boldsymbol{c}_t, \boldsymbol{\alpha}),\quad \boldsymbol{c}_0  = \mathcal{E}(\boldsymbol{\phi}),
\end{equation}
where $\mathcal{F}$ is a neural network that predicts the time derivative of $\boldsymbol{c}$. 
The initial embedding $\boldsymbol{c}_0$ is obtained by encoding the initial condition $\boldsymbol{\phi}$ through $\mathcal{E}$. 
This continuous formulation of latent dynamics allows our model to compute the embeddings at arbitrary time steps through numerical integration and to extrapolate beyond the training horizons.
Our approach departs from previous works~\cite{yin2022continuous,wan2022evolve} by conditioning the prediction of $\mathcal{F}$ on $\boldsymbol{\alpha}$. This empowers the model to effectively capture the rich variety of dynamics governed by different PDE coefficients.

\subsection{Model Training}
\label{sec:optimization}
Most existing EDM methods are predominantly data-driven, relying on extensive datasets of precise solutions for training. However, in many applications, generating sufficient data through repeated simulations or experiments is prohibitively expensive. This problem becomes especially acute in systems with multiple parameters, as the amount of data required scales exponentially with the number of parameters. Such constraints motivate the exploration of physics-informed training, which directly incorporates physical laws into the training process, bypassing the need for large datasets.

\paragraph{Physics-informed loss for EDM}
Given a pair of sampled parameters $\{(\boldsymbol{\phi}^i, \boldsymbol{\alpha}^i)\}$, 
we first train the decoder $\mathcal{D}$ to accurately reconstruct the initial condition $\boldsymbol{\phi}_i$.
To this end, we obtain the initial embedding $\boldsymbol{c}_0^i=\mathcal{E}(\boldsymbol{\phi}_i)$ with auto-decoding given the current decoder $\mathcal{D}$, as defined in~\cref{eq:auto-decoder}. We then optimize $\mathcal{D}$ to reduce the reconstruction error of initial conditions in the auto-decoding process through 
\begin{align}
\begin{split}
\label{eq:nested_ic}
    l_\text{IC}( \boldsymbol{\phi}^i) &\triangleq \mathbb{E}_{\boldsymbol{x}\in\Omega}\|\boldsymbol{\phi}^i(\boldsymbol{x})-\mathcal{D}(\boldsymbol{c}_0^i,\boldsymbol{x})\|_2^2,\, \\
    \text{where}\  \boldsymbol{c}_0^i &= \operatorname*{argmin}_{\boldsymbol{c}} \mathbb{E}_{\boldsymbol{x}\in\Omega}\|\boldsymbol{\phi}^i(\boldsymbol{x}) -\mathcal{D}(\boldsymbol{c}, \boldsymbol{x})\|^2.
\end{split}
\end{align}
While the above loss function involves a nested minimization problem, we employ an efficient approximation in practice.  
Rather than resetting \(\boldsymbol{c}_0^i\) to zero and explicitly solving for the optimal embedding at each step, we assign a learnable embedding for each initial condition at the start of training and update these embeddings, along with the other trainable parameters in the neural networks, using gradient descent. By doing so, we expect \(\boldsymbol{c}_0^i\) to gradually approximate the optimal initial embedding during training, without the need for two-level optimization.
This simplifies~\cref{eq:nested_ic} as:
\begin{equation}
    l_\text{IC}(\boldsymbol{\phi}^i, \mathcal{D}(\boldsymbol{c}_0^i)) \triangleq \mathbb{E}_{\boldsymbol{x}\in\Omega}\|\boldsymbol{\phi}^i(\boldsymbol{x})-\mathcal{D}(\boldsymbol{c}_0^i)(\boldsymbol{x})\|_2^2,
\end{equation}

Then, we leverage the dynamics model $\mathcal{F}$ to unroll the trajectories from initial embedding $\boldsymbol{c}_0^i$. This provides us with the embedding  $\boldsymbol{c}_t^i$ at a sampled time step $t\in[0,T_{tr}]$ and its corresponding prediction $\boldsymbol{\tilde{u}}_t^i(\boldsymbol{x})=\mathcal{D}(\boldsymbol{c}_t^i,\boldsymbol{x})$. Since the exact solution is unavailable, we optimize the PDE residuals instead:
\begin{align}
\begin{split}
\label{eq:pde_residual}
\textstyle 
\frac{\partial\boldsymbol{\tilde{u}}_t^i(\boldsymbol{x})}{\partial t}+\mathcal{L}^{\boldsymbol{\alpha}^i}(\boldsymbol{\tilde{u}}_t^i(\boldsymbol{x})) = \frac{\partial\boldsymbol{\tilde{u}}_t^i(\boldsymbol{x})}{\partial \boldsymbol{c}_t^i}\frac{\partial\boldsymbol{c}_t^i(\boldsymbol{x})}{\partial t}+\mathcal{L}^{\boldsymbol{\alpha}^i}(\boldsymbol{\tilde{u}}_t^i(\boldsymbol{x}))\\
=\frac{\partial\boldsymbol{\tilde{u}}_t^i(\boldsymbol{x})}{\partial \boldsymbol{c}_t^i} \mathcal{F}(\boldsymbol{c}_t, \boldsymbol{\alpha}^i) +\mathcal{L}^{\boldsymbol{\alpha}^i}(\boldsymbol{\tilde{u}}_t^i(\boldsymbol{x})).
\end{split}
\end{align}
Here, the time derivative of $\boldsymbol{c}_t^i(\boldsymbol{x})$ is approximated by $\mathcal{F}$ 
% (with parameters $\boldsymbol{\theta}_{\mathcal{F}}$) 
conditioned on PDE coefficients $\boldsymbol{\alpha}^i$. 
Note that $\boldsymbol{c}_t^i$ is not a trainable embedding. Instead, it's obtained through integration from $\boldsymbol{c}_0^i$ as in \cref{eq:unroll}. Taking altogether, we derive the PDE residual loss as 
\begin{align}
\label{eq:l_pde}
\begin{split}
\textstyle
    l_{\text{PDE}}(\boldsymbol{\alpha}^i, \boldsymbol{c}_t^i, \boldsymbol{\tilde{u}}_t^i) 
    \triangleq \mathbb{E}_{\boldsymbol{x}\in\Omega} \| \frac{\partial\boldsymbol{\tilde{u}}_t^i(\boldsymbol{x})}{\partial \boldsymbol{c}_t^i} \mathcal{F}(\boldsymbol{c}_t, \boldsymbol{\alpha}^i) +\mathcal{L}^{\boldsymbol{\alpha}^i}(\boldsymbol{\tilde{u}}_t^i(\boldsymbol{x}))\|_2^2.
\end{split}
\end{align}
As we parameterize the decoder $\mathcal{D}$ with an INR, the term $\partial\boldsymbol{\tilde{u}}_t^i(\boldsymbol{x})/\partial \boldsymbol{c}_t^i$ and spatial derivatives of $\boldsymbol{\tilde{u}}_t^i(\boldsymbol{x})$ involved in $\mathcal{L}^{\boldsymbol{\alpha}^i}(\boldsymbol{\tilde{u}}_t^i(\boldsymbol{x}))$ can be calculated with AD. 

Similarly, we enforce the boundary conditions with the following loss 
\begin{equation}
    l_{\text{BC}}(\boldsymbol{\tilde{u}}_t^i) \triangleq \mathbb{E}_{\boldsymbol{x}\in\Omega}\|
    \mathcal{B}(\boldsymbol{\tilde{u}}_t^i(\boldsymbol{x}))\|_2^2.
\end{equation}

Overall, the proposed physics-informed training seeks to optimize the parameters of \(\mathcal{D}\) and \(\mathcal{F}\) (denoted as \(\boldsymbol{\theta}_{\mathcal{D}}\) and \(\boldsymbol{\theta}_{\mathcal{F}}\)), along with the embeddings \(\boldsymbol{c}_0^i\) for each training initial condition, to minimize the following objective function
\begin{align}
\label{eq:pi_loss}
\begin{split}
    \min_{(\boldsymbol{\theta}_{\mathcal{F}},\boldsymbol{\theta}_{D}, \boldsymbol{c}_0^i)}  \mathbb{E}_{(\boldsymbol{\phi}^i,\boldsymbol{\alpha}^i)\sim\Phi} \lbrack
    l_\text{IC}(\boldsymbol{\phi}^i, \mathcal{D(\boldsymbol{c_0^i})}) 
    \\
    + \frac{1}{T_{tr}}\sum_{t=1}^{T_{tr}}(l_{\text{PDE}}(\boldsymbol{\alpha}^i, \boldsymbol{c}_t^i, \boldsymbol{\tilde{u}}_t^i)+l_{\text{BC}}(\boldsymbol{\tilde{u}}_t^i))\rbrack. 
\end{split}
\end{align}

\subsection{Diagnosing Physics-informed Optimization in the Latent Space}
\label{sec:diagnose}
\begin{figure*}[t]
% \vskip 0.2in
\begin{center}
\includegraphics[width=0.9\textwidth]{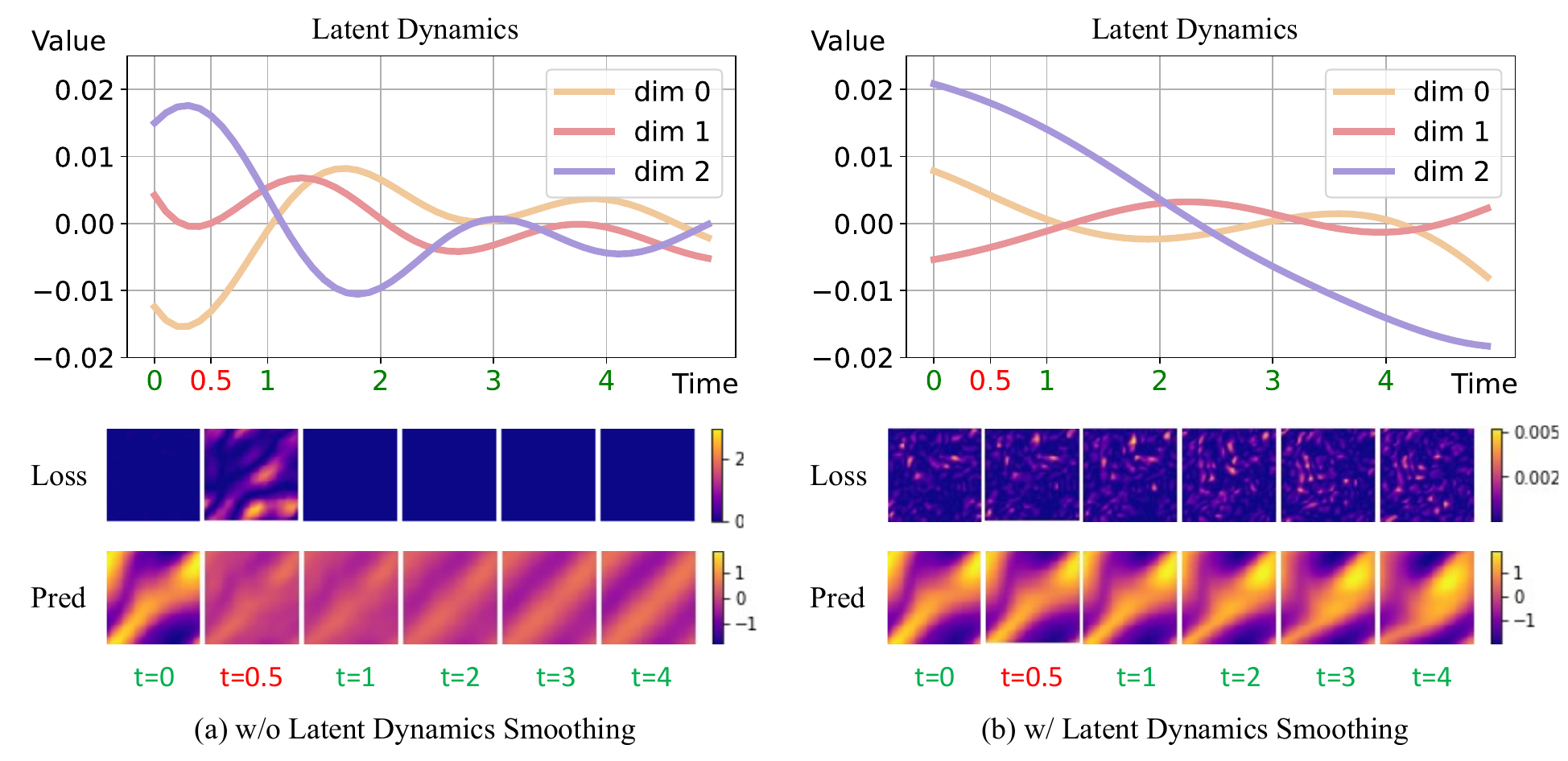}
% \vskip -0.1in
\caption{Training instability arises from overly complex dynamics. We randomly sample 3 dimensions from the 128-dim embeddings. We visualize the loss and predictions of a trajectory at different time steps. Training steps (\(t = \{0, 1, 2, 3, 4\}\)) are highlighted in \textcolor{ForestGreen}{green}, while the test step (\(t = 0.5\)) is shown in \textcolor{red}{red}.}
\label{fig:complex_dyn}
\end{center}
\vskip -0.1in
\end{figure*}

While physics-informed training eliminates the need for exact PDE solutions, it suffers from optimization difficulties~\cite{krishnapriyan2021characterizing, wang2021understanding}, presenting two key challenges in integration with EDM: instability during training and degradation in time extrapolation. Nevertheless, the latent space introduced by EDM offers a novel perspective for diagnosing and addressing these issues. By capturing essential information in low-dimensional representations, it facilitates a more straightforward analysis. Notably, this framework allows us to  identify latent behaviors responsible for each challenge and mitigate them using simple but effective regularization in latent space. 
\subsubsection{Stabilizing the Training Process through Latent Dynamics Smoothing}
In contrast to data-driven methods, physics-informed loss imposes stricter constraints on the time step size to ensure stable training. If the time step is too large, training often collapses into trivial solutions as shown in~\cref{fig:complex_dyn}(a), where the loss is minimized on training steps but remains high on unseen time points close to the initial conditions. Consequently, despite small training losses, the predictions deviate from the ground truth because information fails to propagate from the initial conditions to subsequent steps~\cite{wang2022respecting}. To resolve this, a smaller time step size is often required, which, in turn, significantly increase computational costs and exacerbate the training difficulty.

As an alternative, we explore this issue within the latent space.
Our findings, as illustrated in~\cref{fig:complex_dyn}(a), reveal that the model tends to learn excessively fluctuating latent dynamics from all possible dynamics that minimize training loss. This leads to a complex loss distribution along the time axis, which exhibits poor generalization to unseen points. As a result, the step size constraint becomes even stricter for our method compared to other physics-informed approaches. 
We attribute this to the increased flexibility introduced by the latent dynamics model, which is trained without direct supervision in latent space.
While data-driven EDM methods utilize a similar framework, they do not encounter this issue, as they leverage embeddings from pretrained encoder-decoder networks 
as ground truth for training the dynamics model~\cite{yin2022continuous}. In contrast, our method provides the dynamics model with only indirect supervision through the PDE loss in data space. 

To alleviate this issue, we introduce the \textit{Latent Dynamics Smoothing} regularization, which guides the model to favor simpler dynamics. Inspired by~\cite{finlay2020train}, this regularization mitigates rapid local changing in the predicted trajectories by constraining the time derivative of dynamics model, given by $\frac{\partial \mathcal{F}(\boldsymbol{c}_t, \boldsymbol{\alpha})}{\partial t} = \nabla \mathcal{F}(\boldsymbol{c}_t, \boldsymbol{\alpha}) \cdot \frac{\partial \boldsymbol{c}_t}{\partial t} = \nabla \mathcal{F}(\boldsymbol{c}_t, \boldsymbol{\alpha}) \cdot \mathcal{F}(\boldsymbol{c}_t, \boldsymbol{\alpha}),$
where $\nabla \mathcal{F}(\boldsymbol{c}_t, \boldsymbol{\alpha})$ is the Jacobian matrix of $\mathcal{F}$ with respective to $\boldsymbol{c}_t$. 
Specifically, we apply the latent dynamics smoothing regularization $R_{S}$ to the training time points, where
\begin{equation}
\begin{split}
    R_{S}(\boldsymbol{c}_t, \boldsymbol{\alpha}) &=\|\mathcal{F}(\boldsymbol{c}_t, \boldsymbol{\alpha})\|_{2}^2 + \|\nabla\mathcal{F}(\boldsymbol{c}_t, \boldsymbol{\alpha})\|_{F}^2\\
    &=\|\mathcal{F}(\boldsymbol{c}_t, \boldsymbol{\alpha})\|_{2}^2+\mathbb{E}_{\boldsymbol{\epsilon}\sim\mathcal{N}(0,1)}\|\boldsymbol{\epsilon}^T\nabla\mathcal{F}(\boldsymbol{c}_t, \boldsymbol{\alpha})\boldsymbol{\epsilon}\|_{2}^2.
\end{split}
\end{equation}
As demonstrated in~\cref{fig:complex_dyn}(b), this regularization effectively prevents overly complex dynamics, resulting in a smoother temporal loss distribution without necessitating a reduction in time step size. 

\subsubsection{Improving Time Extrapolation via Latent Dynamics Alignment}

\begin{figure*}[t]
% \vskip 0.2in
\begin{center}
\includegraphics[width=1.0\textwidth]{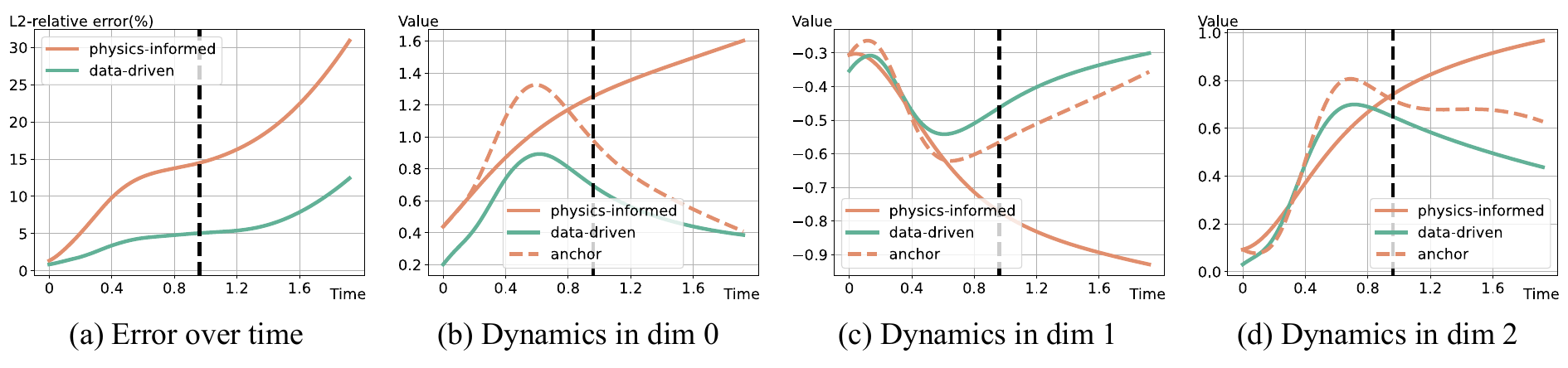}
\vskip -0.1in
\caption{Time extrapolation degradation arises from latent embedding drift. We take the \textbf{CE3} setting as an example. The black dash line separates training and testing horizon. For visualization, we randomly sample 3 dimensions from the 64-dim embeddings.}
\label{fig:drift_dyn}
\end{center}
\vskip -0.1in
\end{figure*}

The next challenge we investigate is the degradation of the model’s time extrapolation ability when trained with physics-informed loss compared to data-driven approaches (\cref{fig:drift_dyn}(a)). 
By visualizing the evolution of latent embeddings over extended time horizons, we identify the issue of latent embedding drift, where the embeddings progressively move outside their typical range as time advances, as depicted in~\cref{fig:drift_dyn}. 
This drift becomes particularly problematic when extrapolating beyond the training horizon, causing the embeddings to deviate from the training distribution, ultimately leading to poor performance at later time steps. 

% \begin{wrapfigure}{R}{0.33\textwidth}
\begin{figure}
% \vskip -0.2in
\begin{center}
\includegraphics[width=0.8\linewidth]{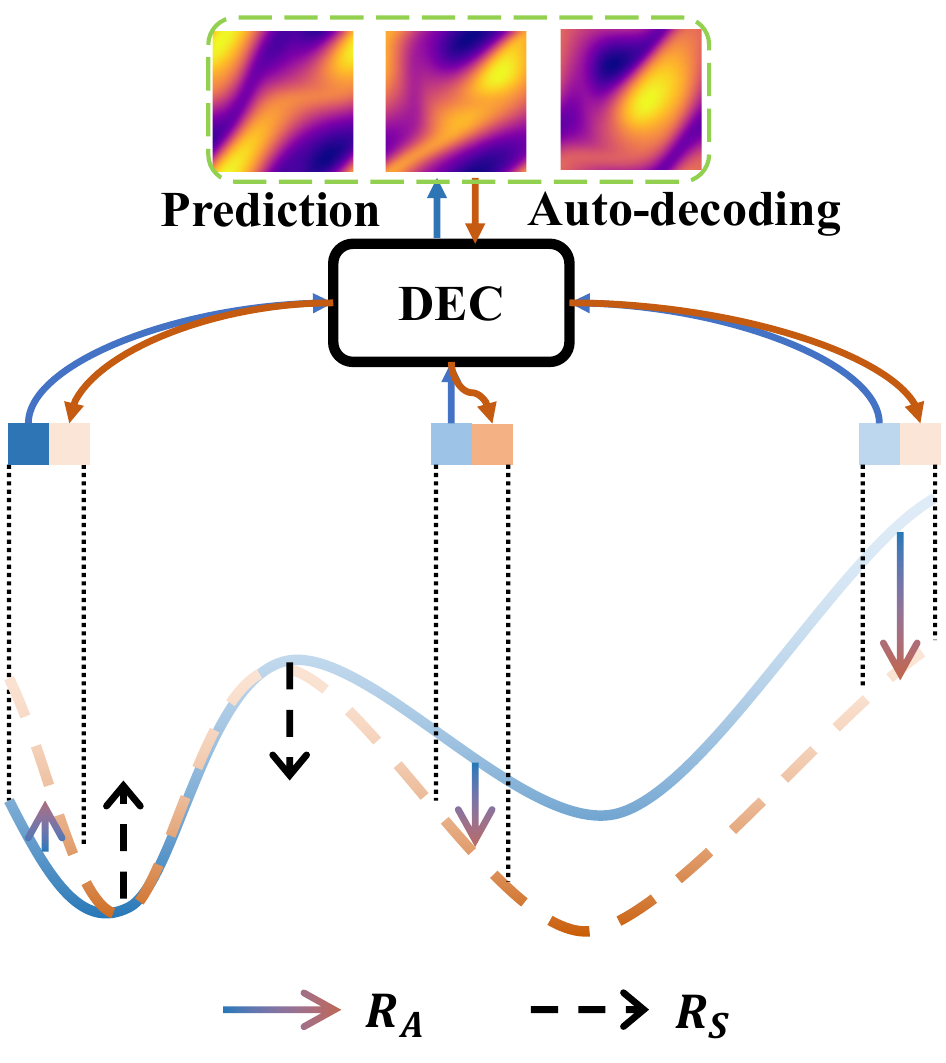}
% \vskip -0.1in
\caption{Regularization. We apply smoothing regularization \(R_{S}\) and alignment regularization \(R_{A}\) to the predicted latent trajectory (the blue curve). \(R_{S}\) prevents rapid local changes, while \(R_{A}\) aligns the predicted latent embeddings with the anchor embeddings (the orange curve), which are obtained by auto-decoding the predicted solutions. This ensures that the embeddings remain within their typical range. }
\label{fig:optimization}
\end{center}
\vskip -0.2in
\end{figure}
We attribute the latent embedding drift to the inconsistency between the supervision signals of the initial embeddings $\boldsymbol{c}_0$ and the later ones $\boldsymbol{c}_t$. To be specific, the initial embeddings are guided by both the exact initial conditions and the PDE loss, whereas the later embeddings rely solely on the PDE loss for supervision. To bridge this gap, we propose to utilize predicted solutions $\boldsymbol{\tilde u}_t=\mathcal{D}(\boldsymbol{c}_t)$ as pseudo labels in the absence of exact PDE solutions. 
While these pseudo labels do not provide additional information in the data space, the embeddings obtained through encoding them, defined as $\boldsymbol{\tilde c}_t=\mathcal{E}(\boldsymbol{\tilde u}_t)=\mathcal{E}(\mathcal{D}(\boldsymbol{c}_t))$, does not exhibit the drift problem as shown in~\cref{fig:drift_dyn}. Thus they can serve as effective anchors to regularize $\boldsymbol{c}_t$ (\cref{fig:optimization}). Note that although $\boldsymbol{\tilde{c}}_t$ and $\boldsymbol{c}_t$ represent the same states $\boldsymbol{\tilde u}_t$, they might not be identical because the neural network is not inherently an one-to-one mapping. The differences arise from their respective training processes. 
Specifically, $\boldsymbol{c}_t$ is unrolled from $\boldsymbol{c}_0$ to satisfy the PDE loss, potentially leading to a distribution shift from that of $\boldsymbol{c}_0$. 
In contrast, $\boldsymbol{\tilde{c}}_t$ are obtained via auto-decoding from the data space (similar to $\boldsymbol{c}_0$), resulting in a distribution closer to the initial embeddings. Therefore, we can mitigate the drift problem by aligning latent embeddings $\boldsymbol{c}_t$ with anchor embeddings $\boldsymbol{\tilde c}_t$ via the \textit{Latent Dynamics Alignment} regularization $R_{A}(\boldsymbol{c}_t,\boldsymbol{\tilde{c}}_t) = \|\boldsymbol{c}_t-\boldsymbol{\tilde{c}}_t \|_2$.
Note that this alignment is achieved with minimal impact on the predicted solution, as both embeddings correspond to the same output.

\begin{algorithm}[tb]
   \caption{Training Procedure}
   \label{alg:train}
   \definecolor{codeblue}{rgb}{0.25,0.5,0.5}
    \renewcommand{\algorithmiccomment}[2][.45\linewidth]{%
  \leavevmode\hfill\makebox[#1][l]{//~#2}}
\begin{algorithmic}
   \STATE {\bfseries Input:} Parameters $\boldsymbol{\theta}_{D}$ and $\boldsymbol{\theta}_{\mathcal{F}}$, initial conditions and PDE coefficients $(\boldsymbol{\phi}^i,\boldsymbol{\alpha}^i)$, initial embeddings $(\boldsymbol{c}_0^i)$, latent embeddings for consistency regularization $(\{\boldsymbol{\tilde{c}}_t^i\}_{t=1}^{T_{tr}})$
   \STATE {\bfseries Initialize:} $\boldsymbol{c}_0^i\leftarrow 0, \forall i$ and $\boldsymbol{\tilde{c}}_t^i\leftarrow 0, \forall (i,t)$
   \STATE {\bfseries Hyper-parameters:} learning rates $\lambda_{c}$ for latent embeddings, $\lambda_{D}$ for $\boldsymbol{\theta}_{D}$  and $\lambda_{\mathcal{F}}$ for $\boldsymbol{\theta}_{\mathcal{F}}$;
   \REPEAT
   \STATE Sample one pair of $(\boldsymbol{\phi}^i,\boldsymbol{\alpha}^i, \boldsymbol{c}_0^i, \{\boldsymbol{\bar{c}}_t^i\}_{t=1}^{T_{tr}})$;
   \STATE \textcolor{codeblue}{$/*$\ unroll the latent trajectory \ $*/$}
   \STATE $\{\boldsymbol{c}_t^i\}_{t=1}^{T_{tr}} \leftarrow \mathcal{F}_{\boldsymbol{\theta}_{\mathcal{F}}}(\boldsymbol{c}_0^i, \boldsymbol{\alpha}^i)$;    
   \STATE \textcolor{codeblue}{$/*$\ obtain predicted solutions \ $*/$}
   \STATE $\boldsymbol{\tilde{u}}_t^i \leftarrow \mathcal{D}_{\boldsymbol{\theta}_{D}}(\boldsymbol{c}_t^i), \ t=0,1,...,T_{tr}$; 
   %\textcolor{codeblue}{\COMMENT{obtain predicted solutions}}
   \STATE \textcolor{codeblue}{$/*$\ update anchor embeddings for alignment regularization $R_{A}$\ $*/$}
    
    \STATE $\boldsymbol{\tilde{c}}_t^i \leftarrow \boldsymbol{\tilde{c}}_t^i-\lambda_c \nabla_{\boldsymbol{\tilde{c}}_t^i} \mathbb{E}_{\boldsymbol{x}\in\Omega}\|\mathcal{D}_{\boldsymbol{\theta}_{D}}(\boldsymbol{\tilde{c}}_t^i)(\boldsymbol{x})-\boldsymbol{\tilde{u}}_t^i(\boldsymbol{x})\|_2^2, \forall t$ ; 
    %\textcolor{codeblue}{\COMMENT{prepare for the auto-decoding of predicted solutions}}
   \STATE \textcolor{codeblue}{$/*$\ update initial embeddings \ $*/$}
    \STATE $\boldsymbol{c}_0^i \leftarrow \boldsymbol{c}_0^i-\lambda_c \nabla_{\boldsymbol{c}_0^i} l_\text{IC}(\boldsymbol{\phi}_i, \boldsymbol{\tilde{u}}_0^i)$; % \textcolor{blue}{\COMMENT{update initial embeddings}}
     \STATE \textcolor{codeblue}{$/*$\ update network parameters with physics-informed loss and regularization\ $*/$}
    \STATE $\boldsymbol{\theta}_{D} \leftarrow \boldsymbol{\theta}_{D}-\lambda_D \nabla_{\boldsymbol{\theta}_{D}} \lbrack l_\text{IC}(\boldsymbol{\phi}_i, \boldsymbol{\tilde{u}}_0^i) + \frac{1}{T_{tr}}\sum_{t=1}^{T_{tr}}(l_{\text{BC}}(\boldsymbol{\tilde{u}}_t^i))+l_{\text{PDE}}(\boldsymbol{\alpha}^i, \boldsymbol{c}_t^i, \boldsymbol{\tilde{u}}_t^i)\rbrack$; %\textcolor{blue}{\COMMENT{update initial embeddings}}
   
    \STATE $\boldsymbol{\theta}_{\mathcal{F}} \leftarrow \boldsymbol{\theta}_{\mathcal{F}}-\lambda_{\mathcal{F}} \nabla_{\boldsymbol{\theta}_{\mathcal{F}}} \lbrack l_\text{IC}(\boldsymbol{\phi}_i, \boldsymbol{\tilde{u}}_0^i) + \frac{1}{T_{tr}}\sum_{t=1}^{T_{tr}}(l_{\text{BC}}(\boldsymbol{\tilde{u}}_t^i)+l_{\text{PDE}}(\boldsymbol{\alpha}^i, \boldsymbol{c}_t^i, \boldsymbol{\tilde{u}}_t^i)+R_{S}(\boldsymbol{c}^i_t, \boldsymbol{\alpha}_i)+R_{A}(\boldsymbol{c}^i_t,\boldsymbol{\tilde{c}}^i_t))\rbrack$; %\textcolor{blue}{\COMMENT{update initial embeddings}}

   \UNTIL{convergence}
\end{algorithmic}
\end{algorithm}

\begin{algorithm}[tb]
   \caption{Testing Procedure}
   \label{alg:test}
   \definecolor{codeblue}{rgb}{0.25,0.5,0.5}
    \renewcommand{\algorithmiccomment}[2][.6\linewidth]{%
  \leavevmode\hfill\makebox[#1][l]{//~#2}}
\begin{algorithmic}
   \STATE {\bfseries Input:} Parameters $\boldsymbol{\theta}_{D}$ and $\boldsymbol{\theta}_{\mathcal{F}}$, initial conditions and PDE coefficients $(\boldsymbol{\phi},\boldsymbol{\alpha})$, initial embeddings $(\boldsymbol{c}_0)$
   \STATE {\bfseries Initialize:} set embeddings to zero $\boldsymbol{c}_0\leftarrow 0$
   \STATE {\bfseries Hyper-parameters:} learning rates $\lambda_{c}$ for latent embeddings, optimization step for auto-decoding $S$;
    \STATE \textcolor{codeblue}{$/*$\ Update $\boldsymbol{c}_0$ with auto-decoding\ $*/$}
   \FOR{$s=1$ {\bfseries to} $S$}
    \STATE $\boldsymbol{c}_0 \leftarrow \boldsymbol{c}_0-\lambda_c \nabla_{\boldsymbol{c}_0} \mathbb{E}_{\boldsymbol{x}\in\Omega}\|\mathcal{D}_{\boldsymbol{\theta}_{D}}(\boldsymbol{c}_0)(\boldsymbol{x})-\boldsymbol{\phi}(\boldsymbol{x})\|_2^2$;
   \ENDFOR
   \STATE \textcolor{codeblue}{$/*$\ unroll the latent trajectory \ $*/$}
   \STATE $\{\boldsymbol{c}_t\}_{t=1}^{T_{te}} \leftarrow \mathcal{F}_{\boldsymbol{\theta}_{\mathcal{F}}}(\boldsymbol{c}_0, \boldsymbol{\alpha})$;    
   \STATE \textcolor{codeblue}{$/*$\ obtain predicted solutions \ $*/$}
   \STATE $\boldsymbol{\tilde{u}}_t \leftarrow \mathcal{D}_{\boldsymbol{\theta}_{D}}(\boldsymbol{c}_t), \ t=0,1,...,T_{te}$.
\end{algorithmic}
\end{algorithm}

\section{Experiments}
\label{sec:experiments}
We begin by introducing the benchmarks in~\cref{sec:benchmark}. \cref{sec:main_results} presents the main results of our study. Finally, \cref{sec:downstream} explores the transferability of the pre-trained {\name} to downstream tasks. 

\subsection{Benchmarks}
\label{sec:benchmark}

\paragraph{1D combined equations}
We consider the family of PDEs:
\begin{equation}
\label{eq:ce}
    \begin{split}
    \frac{\partial u}{\partial t}+2u\frac{\partial u}{\partial x}-\alpha_0\frac{\partial^2u}{\partial x^2} +\alpha_1\frac{\partial^3u}{\partial x^3}=0,\quad  u(t=0,x)=\phi(x),
\end{split}
\end{equation}
where $x\in[0,L_x]$ and $t\in[0,T]$. 
This equation 
encompasses several fundamental physical phenomena, namely nonlinear advection, viscosity, and dispersion. 
It is a combination of Burgers' equation (when $\alpha_1=0$) and Korteweg–De Vries (KdV) equation (when $\alpha_0=0$).  
We aim to predict $u(t,x)$ given varying $\phi(x)$ and $\boldsymbol{\alpha}=(\alpha_0,\alpha_1)$.  
We consider three scenarios: $\bullet$ \textbf{CE1} for Burgers' equation with $\boldsymbol{\alpha}=(0.1,0)$; $\bullet$ \textbf{CE2} for KdV equation with $\boldsymbol{\alpha}=(0,0.05)$; and $\bullet$ \textbf{CE3} for combined equation with $\boldsymbol{\alpha}\in\{(\alpha_0, \alpha_1)|0<\alpha_0\leq0.4,0<\alpha_1\leq0.65\}$. More details can be found in Appendix~\ref{sec:training_details}.

\paragraph{2D Navier-Stokes equations} We consider the 2D Navier-Stokes equations, which describe the dynamics of a viscous and incompressible fluid. The equations are given by 
\begin{equation}
\label{eq:ns}
\begin{split}
&\frac{\partial w}{\partial t} + \boldsymbol{u} \cdot \nabla w - \frac{1}{\alpha}\Delta w-f =0,\\
&w = \nabla \times \boldsymbol{u},\quad \nabla \cdot \boldsymbol{u} = 0,\quad w(t=0,\boldsymbol{x})=\phi(\boldsymbol{x}),
\end{split}
\end{equation}
where $\boldsymbol{x}\in[0,1]^2$, $w$ is vorticity, $\boldsymbol{u}$ is the velocity field and f is the forcing function. 
We consider the long temporal transient flow with the forcing term $f(\boldsymbol{x})=0.1(\mathrm{sin}(2\pi(x_1+x_2))+\mathrm{cos}(2\pi(x_1+x_2)))$ following previous works~\cite{li2020fourier,yin2022continuous}.
The Reynolds number, denoted as $\alpha$, serves as an indicator of the fluid viscosity. The modeling of fluid dynamics becomes increasingly challenging with higher Reynolds numbers, since the flow patterns transition to complex and chaotic regimes.
We investigate the prediction of vorticity dynamics under varying initial conditions $\phi$ and Reynolds numbers $\alpha$ with two scenarios: $\bullet$ \textbf{NS1} focuses on a fixed $\alpha=1000$, 
while $\bullet$ \textbf{NS2} considers varying Reynolds numbers with $\alpha\in [700,\ 1400]$.
See Appendix~\ref{sec:training_details} for more details.

\paragraph{Tasks and metrics} For each problem, we create training and test sets by randomly sampling $N_{tr}$ and $N_{ts}$ parameter-solution pairs $\{(\boldsymbol{\phi}_i,\boldsymbol{\alpha}_i, \boldsymbol{u}_i)\}$, respectively. The exact solution $\boldsymbol{u}_i$ is only used for evaluation.
For each prediction $\boldsymbol{\hat{u}}_i$, we assess the $L_2$ relative error, expressed as $\|\boldsymbol{\hat{u}}_i-\boldsymbol{u}_i\|_2 / \|\boldsymbol{u}_i\|_2$, over the full time interval $[0, T]$, subdivided into the training horizon $[0, T_{tr})$ (denoted as \textit{In-t}) and the subsequent duration $[T_{tr}, T]$ (denoted as \textit{Out-t}). 

\subsection{Main results} 
\label{sec:main_results}

\begin{table*}[t]
\caption{Results on the test set of 1D and 2D benchmarks. We report the $L_2$ relative error (\%) over the training horizon (\textsc{In-t}) and the subsequent duration (\textsc{Out-t}). The best results are \textbf{bold-faced}.}
\vskip -0.2in
\label{exp:ce}
\begin{center}
\begin{small}
\begin{sc}
% \resizebox{1.0\textwidth}{!}{
\begin{tabular}{lcccccccccc}
\toprule
 & \multicolumn{2}{c}{CE1} & \multicolumn{2}{c}{CE2} & \multicolumn{2}{c}{CE3} & \multicolumn{2}{c}{NS1} & \multicolumn{2}{c}{NS2} \\ 
 \cmidrule{2-3} \cmidrule{4-5} \cmidrule{6-7}  \cmidrule{8-9} \cmidrule{10-11}  
Model & In-t & Out-t & In-t & Out-t & In-t & Out-t & In-t & Out-t & In-t & Out-t\\
\midrule
PI-DeepONet & 4.18  & 8.61 & 17.17 & 36.16 & 7.57 & 15.74 & 14.80& 21.19 & 16.31 & 25.41\\
PINODE & 10.44  & 24.75 & 11.03 & 28.69 & 18.21 & 39.41 & 16.44 & 53.52 & 17.56 & 46.67\\
MAD & 3.98 & 9.32 & 12.00 & 27.97 & 6.78 & 17.10 & 9.83& 19.92& 11.23& 21.87\\
{\name} & \textbf{1.48} & \textbf{2.24} & \textbf{3.02} & \textbf{7.15} & \textbf{3.19} & \textbf{8.08} & \textbf{2.35} & \textbf{5.43} & \textbf{4.59} & \textbf{10.02}\\
\bottomrule
\end{tabular}
\end{sc}
\end{small}
\end{center}
\vskip -0.1in
\end{table*}

\begin{figure*}[t]
% \vskip 0.2in
\begin{center}
\includegraphics[width=1.0\textwidth]{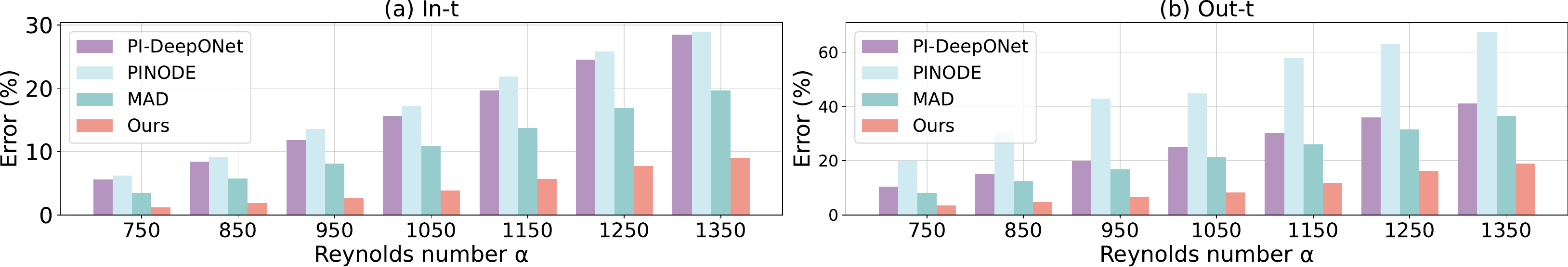}
% \vskip -0.1in
\caption{Performance (y-axis) of various models on the \textbf{NS2} test set with unseen Reynolds numbers (x-axis). $L_2$ relative error ($\%$) is reported.}
\label{fig:reynolds_number}
\end{center}
% \vskip -0.3in
\end{figure*}

We compare {\name} with PI-DeepONet~\cite{wang2021learning}, PINODE~\cite{sholokhov2023physics} and MAD~\cite{huang2022meta} on the benchmarks detailed in~\cref{sec:benchmark} (see implementation details in Appendix~\ref{sec:implment} and computational efficiency in Appendix~\ref{sec:com_eff}). 
For PINODE, which requires input data sampled from the exact solution distribution (unavailable here), we use the distribution of initial conditions as a substitute because these are the only data accessible (see Appendix~\ref{sec:detail_auto_encoder} for details). 
Performance of all methods on the training and test set is reported in~\cref{exp:ce}. 
Additional results (training set performance, sample efficiency analysis and visualizations) are available in Appendix~\ref{sec:add_re}.

\textbf{Generalizing across initial conditions.} Focusing on scenarios with fixed coefficients, our method achieves the lowest error in test set \textit{In-t}. Specifically, it outperforms the second-best method by substantial margins: 63\% for \textbf{CE1} (1.48\% vs. 3.98\%), 72\% for \textbf{CE2} (3.02\% vs. 11.03\%), and 76\% for \textbf{NS1} (2.35\% vs. 9.83\%). These results highlight {\name}'s effectiveness in handling diverse initial conditions within these benchmarks.

\textbf{Generalizing across PDE coefficients.} 
We evaluate the models' ability to handle unseen PDE coefficients in scenarios like \textbf{CE3} and \textbf{NS2}. For PI-DeepONet and PINODE, we provide coefficients $\boldsymbol{\alpha}$ as additional inputs alongside the initial conditions. This allows them to make predictions conditioned on both factors.
In this setting, our {\name} demonstrates remarkable robustness to changes in PDE coefficients. \cref{fig:reynolds_number} presents the relative error for various test Reynolds numbers in \textbf{NS2} scenario. We observe that as the Reynolds numbers increase, {\name} consistently maintains a low solution error. 
See Appendix~\ref{sec:extra_alpha} for {\name}'s extrapolation ability beyond the training distribution.

\textbf{Generalizing beyond training horizon.} To assess the models' capability to predict past the training horizon, we adopt the auto-regressive evaluation strategy from \cite{wang2023long} for PI-DeepONet and MAD. 
This approach iteratively extends forecasts by using the final state predicted in the training interval as the initial condition for the next one. 
Across all scenarios, {\name} demonstrates clear superiority over the baseline methods on \textit{Out-t}. 
While PINODE utilizes a Neural ODE capable of integrating beyond the training horizon, its performance suffers in practice. We hypothesize this limitation stems from its input data distribution failing to capture the true distribution of the solutions.

% \begin{wraptable}{r}{0.5\textwidth}
\begin{table}[t]
% \vskip -0.15in
\caption{Comparison with DIN\textsc{o} trained with different sub-sampling ratios of training set. $L_2$ relative error in \textbf{NS1} is reported (\%). }
\label{exp:data_driven}
% \vskip -0.2in
\begin{center}
\begin{small}
\begin{sc}
\resizebox{0.9\linewidth}{!}{
\begin{tabular}{lcccccc}
\toprule
 & & \multicolumn{2}{c}{Train} & & \multicolumn{2}{c}{Test} \\ \cmidrule{3-4} \cmidrule{6-7} 
Model & Dataset & In-t & Out-t & & In-t & Out-t \\
\midrule
\name & - & 1.81 & 4.10 & & \textbf{2.35} & \textbf{5.43} \\
DIN\textsc{o} & 100\% & \textbf{1.42} & \textbf{3.32} & & 4.26 & 5.73\\
\midrule
DIN\textsc{o} & 50\% & 1.14 & 4.90 & & 5.25 & 8.76 \\
DIN\textsc{o} & 25\% & 0.82 & 7.52 & & 7.37 & 13.58\\
DIN\textsc{o} & 12.5\% & 0.47 & 13.59 & & 15.12 & 31.74 \\
\bottomrule
\end{tabular}}
\end{sc}
\end{small}
\end{center}
% \vskip -0.1in
% \end{wraptable}
\end{table}

\textbf{Comparison with the data-driven approach.} We compare {\name} with the data-driven counterpart, DIN\textsc{o}~\cite{yin2022continuous}, in~\cref{exp:data_driven}. 
Despite achieving a slightly higher training error than DIN\textsc{o} trained with the full dataset, {\name} outperforms it on the test set, demonstrating a superior ability to adapt to unseen initial conditions. 
We attribute this improvement to the incorporating with physics-informed training, which ensures our model produces physically plausible predictions and reduces overfitting to noise and anomalies in the data. 
Furthermore, DIN\textsc{o}'s performance degrades with smaller training sets, emphasizing the limitations of data-driven approaches in achieving robust generalization with limited data. See Appendix~\ref{sec:more_data_driven} for comparisons with more data-driven baselines.

% \begin{wraptable}{r}{0.5\textwidth}
\begin{table}[t]
\caption{Ablations on regularization. We report $L_2$ relative error (\%) within $i$-th time interval $[(i-1)\Delta T$, $i\Delta T)$, where $i\in\{1,2,3,4\}$ and $\Delta T=T_{tr}$. The default setting marked in \colorbox{gray!25}{gray}. }
% \vskip -0.15in
\label{exp:reg}
% \vskip 0.15in
\begin{center}
\begin{small}
\begin{sc}
\resizebox{0.9\linewidth}{!}{
\begin{tabular}{cc|cccc}
\toprule
$R_{A}$ & $R_{S}$ & 1st$\Delta T$ & 2nd$\Delta T$ & 3rd$\Delta T$ & 4th$\Delta T$ \\
\midrule
\rowcolor{gray!25}
\cmark & \cmark & \textbf{4.59} & \textbf{10.02} & \textbf{14.90} & \textbf{19.57} \\
\xmark & \cmark & 4.76 & 11.00 & 17.02 & 24.53 \\ \
\xmark & \xmark & 17.90 & 33.83 & 43.98 & 51.87 \\ 
\bottomrule
\end{tabular}}
\end{sc}
\end{small}
\end{center}
\vskip -0.2in
% \end{wraptable}
\end{table}
\textbf{Ablations on regularization methods.} We provide ablation studies to verify the effectiveness of two regularization methods. We present the results in the test set of \textbf{NS2} scenario with inference extended to four times the training interval. As depicted in \cref{exp:reg}, we can find that the unregularized physics-informed training fails to deliver accurate prediction, underscoring the indispensability of all regularization methods for achieving optimal performance. Removing the alignment regularization $R_{A}$ leads to a significant drop in performance on long-range prediction, demonstrating its crucial contribution to the model's temporal extrapolation ability.  
Notably, without $R_{S}$, the model struggles to converge to an acceptable level of performance, highlighting the necessity of learning simple latent dynamics for  stable and effective training.

\subsection{Downstream Tasks}
\label{sec:downstream}
Having established {\name}'s robustness to unseen initial conditions and PDE coefficients, we now explore its representation transferability to downstream tasks. 
Ideally, transferable representations should simplify learning for subsequent problems.
Here, we investigate how a {\name} pre-trained on the \textbf{NS2} scenario performs on downstream tasks like long-term integration and inverse problems.

\begin{table*}[t]
\caption{Downstream tasks on NS equations. For long-term integration, we report the accumulated $L_2$ relative error (\%) in $[0,i\Delta T)$, where $i\in\{1,4,7,10\}$ and $\Delta T=T_{tr}$. For inverse problem, we report the $L_2$ relative error (\%) of the predicted PDE coefficients under different snapshots N.
We compare different training settings of {\name}, including training from scratch (FS), finetuning the dynamics model of pretrained {\name} (FT-DYN) and finetuning its all components (FT-ALL).} 
\subfloat[Long-term integration]{
% \label{exp:inverse}
% \vskip 0.15in
% \begin{center}
% \begin{small}
% \begin{sc}
% \caption{Long-term integration}
% \vskip -0.1in
\label{exp:longterm}
\resizebox{0.53\textwidth}{!}{
\begin{tabular}{ccccccc}
\toprule
 & \multicolumn{2}{c}{PI-DeepONet} & &\multicolumn{3}{c}{\name} \\
 \cmidrule{2-3} \cmidrule{5-7}
& FS & FT & & FS & FT-DYN & FT-ALL\\
\midrule
1$\Delta T$ & 8.85& 5.95 & & 1.01 & 1.52 & \textbf{0.53} \\
4$\Delta T$ & 57.85& 20.31 & & 13.50 & 5.15 & \textbf{2.38} \\
7$\Delta T$ & 64.14& 26.11 & & 29.11 & 7.71 & \textbf{3.99} \\
10$\Delta T$ &68.45&24.68 & & 36.53 & 8.33 & \textbf{5.75} \\
\bottomrule
\end{tabular}}
}
% \end{sc}
% \end{small}
% \end{center}
% \end{subtable}
% \hfill
% \begin{subtable}[h]{0.43\textwidth}
\subfloat[Inverse problem]{
% \vskip 0.15in
% \begin{center}
% \begin{small}
% \begin{sc}
% \caption{Inverse problem}
% \vskip -0.1in
\label{exp:inverse}
\resizebox{0.43\textwidth}{!}{
\begin{tabular}{lcccc}
\toprule
 & PINN & \multicolumn{3}{c}{\name} \\
 \cmidrule{3-5}
 & FS & FS & FT-DYN & FT-ALL\\
\midrule
N=10 & 2.69  & 1.34& 0.17 & \textbf{0.07} \\
N=5 & 3.38  & 2.22 & 0.31 & \textbf{0.21} \\
N=3 & 8.99  & 9.54 & 2.44 & \textbf{1.80} \\
N=2 & 15.47  & 14.39 & 3.63 & \textbf{2.92} \\
\bottomrule
\end{tabular}}
% \end{sc}
% \end{small}
% \end{center}
% \end{subtable}
}
% \vskip -0.2in
\end{table*}

\paragraph{Long-term Integration}
Training a physics-informed neural PDE solver for long temporal horizons presents a significant challenge. 
We adopt the training setting outlined in \cite{wang2023long}, which reformulates the long-term integration problem as a series of initial value problems solved within a shorter horizon ($T_{tr}$). 
Data snapshots are uniformly sampled across the entire test horizon ($T_{ts}$) to serve as training initial conditions.
The model, trained on the shorter interval $T_{tr}$, can iteratively predict future states by using previous prediction at the end of $T_{tr}$ as new initial condition, extending its effective horizon without direct long-term training. While this approach can be sensitive to the number of training initial conditions, {\name}'s pre-training knowledge mitigate this limitation. 

\begin{figure*}[t]
% \vskip 0.2in
\begin{center}
\centerline{\includegraphics[width=0.90\textwidth]{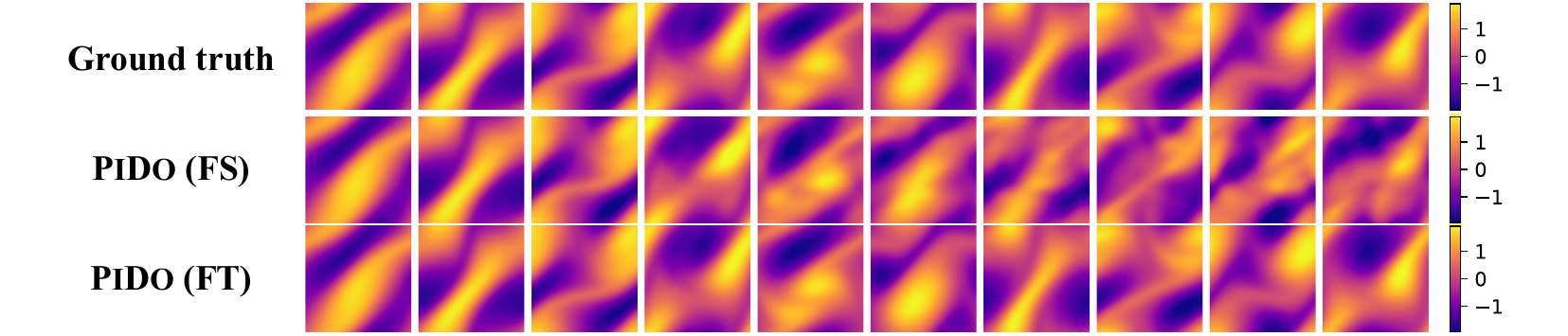}}
% \vskip -0.2in
\caption{Visualization of {\name}'s predictions in the long-term integration setting, spanning from \(t = 0\) s to \(t = 45\) s with a step size of 5 s.}
\label{fig:longterm_prediction}
\end{center}
\vskip -0.2in
\end{figure*}

We consider a challenging setting where the test horizon $T_{ts}$ is ten times longer than the training horizon $T_{tr}$. We set $T_{tr}$ to 5s. The data is generated with a Reynolds number $\alpha=950$, which is unseen by {\name} during the pre-training stage. 
Ten data snapshots are uniformly sampled from the test horizon as training initial conditions.
We compare {\name} against PI-DeepONet. When inference, {\name} can directly predict the entire horizon, while PI-DeepONet relies on the iterative scheme. 
As shown in \cref{exp:longterm}, both PI-DeepONet and {\name} struggle to achieve satisfactory long-term performances when trained from scratch. 
In contrast, the pre-trained {\name} exhibits significant improvement (77\% error reduction) through fine-tuning the dynamics model. This showcases its effectiveness in enhancing long-term predictions with insufficient data. Furthermore, fine-tuning all components of the pre-trained {\name} can lead to further performance gains. We provide in~\cref{fig:longterm_prediction} visualizations of {\name}\ (FS) and {\name}\ (FT) in the long-term integration task.

% \subsubsection{Inverse Problem}
\paragraph{Inverse Problem} 
PINNs have demonstrated efficacy in solving inverse problems~\cite{raissi2019physics,raissi2020hidden}, aiming to recover the PDE coefficients $\boldsymbol{\alpha}$ from a limited set of observations~\cite{pakravan2021solving,zhao2022learning,nair2023grids}.
We then showcase the effectiveness of {\name}'s pretrained knowledge in this context by comparing it against PINN. 
We follow \cite{raissi2019physics} to treat coefficients $\boldsymbol{\alpha}$ as learnable parameters and optimize them with a neural network to simultaneously fit the observed data and satisfy the PDE constraints.
We focus on a case with Reynolds number $\alpha=950$ and a training horizon of $T_{tr}=10s$. Training data consists of N solution snapshots uniformly sampled across the entire horizon. For each snapshot, 5\% of spatial locations are randomly selected as observed data points. 
\cref{exp:inverse} shows that {\name} consistently outperforms its from-scratch counterpart when only finetuning the dynamics model. Notably, even with only two snapshots, the pretrained {\name} achieves accurate predictions.
This highlights the capability of {\name}'s transferable representations in alleviating the data scarcity burden in inverse problems.

\section{Discussion and Conclusion}
\label{sec:conclusion}
In this paper, we propose {\name}, a novel physics-informed neural PDE solver demonstrating exceptional generalization across diverse PDE configurations. 
{\name} effectively leverages the shared structure of dynamical systems by projecting solutions into a latent space and learning their dynamics conditioned on PDE coefficients. 
To tackle the challenges of physics-informed dynamics modeling, we adopt an innovative perspective by diagnosing and mitigating them in latent space, resulting in a significant improvement in the model's temporal extrapolation and training stability.
Extensive experiments on 1D and 2D benchmarks demonstrate {\name}'s generalization ability to initial conditions, PDE coefficients and training time horizons, along with transferability to downstream tasks. 

We acknowledge two key limitations that motivate future research directions:
First, while we employ periodic boundary conditions in all our experiments, a crucial future direction is to extend the generalization study to handle diverse boundary conditions and geometries.  A promising approach might involve utilizing multiple decoders, each tailored to specific boundary conditions.
Second, the smoothing regularization used for stability training can potentially penalize high-frequency  information, which is crucial for accurate modeling. Currently, we achieve a trade-off between stability and accuracy with a suitable regularization weight. It is a promising avenue to investigate alternative regularization that improves stability without compromising the capture of high-frequency details. 
% \section*{Acknowledgments}
% The work is supported by the National Natural Science Foundation of China under Grants 62276150.

% \newpage
\appendices
% \onecolumn
\section{Additional Results}
\label{sec:add_re}

\subsection{Extrapolation outside the Training Distribution of PDE Coefficients. }
\label{sec:extra_alpha}
We examine the extrapolation capability of {\name} beyond the training distribution of Reynolds numbers in the \textbf{NS2} setting. We focused on higher Reynolds numbers as they represent more complex fluid dynamics. Our comparisons with MAD, the best-performing baseline, are detailed in~\cref{exp:extra_alpha}.
\begin{table*}[ht]
% \vskip -0.2in
\caption{Extrapolation outside the training distribution of Reynolds number in \textbf{NS2} setting. We report the $L_2$ relative error (\%) over the training horizon (\textsc{In-t}) and the subsequent duration (\textsc{Out-t}).}
\label{exp:extra_alpha}
% \vskip 0.15in
\begin{center}
\begin{small}
\begin{sc}
\begin{tabular}{lccccccc}
\toprule
Model & $\alpha=550$ & $\alpha=650$ & $\alpha=1450$ & $\alpha=1550$ & $\alpha=1650$ & $\alpha=1750$ & $\alpha=1850$ \\
\midrule
\textit{In-t} \\
\midrule 
MAD & 1.83 & 2.27   &20.73 &22.45 &25.09 &27.46  &29.61\\
{\name} & 1.08 & 0.68 & 10.77 & 12.17 & 13.89 & 15.04 & 16.51 \\
\midrule
\textit{Out-t} \\
\midrule 
MAD & 3.71 & 5.30&  38.70 &43.07  &48.57 &52.78  &56.15\\
{\name} & 3.38 & 2.11 & 22.32 & 25.90 & 30.29 & 32.90 & 36.60 \\
\bottomrule
\end{tabular}
\end{sc}
\end{small}
\end{center}
\vskip -0.2in
\end{table*}

\subsection{Comparisons with Data-driven Baselines}
\label{sec:more_data_driven}
In addition to DIN\textsc{o}, we compare our method with other data-driven approaches, including FNO~\cite{li2020fourier} and DeepONet~\cite{lu2021learning}, in \cref{exp:more_data_driven}. Our results show that {\name} consistently outperforms both FNO and DeepONet in the \textit{In-t} and \textit{Out-t} settings.

We also compare our method with PINO~\cite{li2021physics}, which approximates the PDE-based loss using the finite difference method, making it sensitive to the time step size. To ensure training stability, we adopt a time step size of 0.2 seconds for PINO, which is five times smaller than that used for {\name}. Our results indicate that PIDo demonstrates competitive performance with PINO in the \textit{In-t} prediction but outperforms it in the \textit{Out-t} scenario.
\begin{table}[ht]
% \vskip -0.15in
\caption{Comparison with data-driven methods. $L_2$ relative error in \textbf{NS1} is reported (\%). }
\label{exp:more_data_driven}
% \vskip -0.2in
\begin{center}
\begin{small}
\begin{sc}
\resizebox{1.0\linewidth}{!}{
\begin{tabular}{lcccccc}
\toprule
 & & \multicolumn{2}{c}{Train} & & \multicolumn{2}{c}{Test} \\ \cmidrule{3-4} \cmidrule{6-7} 
Model & Dataset & In-t & Out-t & & In-t & Out-t \\
\midrule
PINO & - & 3.88 & 10.11 & & 3.89 & 10.16 \\
\name & - & 1.81 & 4.10 & & \textbf{2.35} & \textbf{5.43} \\
\midrule
DeepONet & 100\% & 7.15 & 12.23 & & 10.28 & 13.55 \\
FNO & 100\% & 2.83 & 8.74 & & 2.86 & 8.83 \\
DIN\textsc{o} & 100\% & \textbf{1.42} & \textbf{3.32} & & 4.26 & 5.73\\
\bottomrule
\end{tabular}}
\end{sc}
\end{small}
\end{center}
\vskip -0.1in
\end{table}

\subsection{Ablation on INR Architectures}
We investigate the impact of different choices for the INR architectures within PI-DeepONet, MAD, and our proposed method in~\cref{exp:ablation_inr}. The results demonstrate that employing FourierNets consistently improves the performance of all three methods on both \textit{In-t} and \textit{Out-t} metrics compared to using MLPs with tanh or sin activation functions.  Furthermore, our method achieves superior performance over both PI-DeepONet and MAD in all evaluated INR settings. 
\begin{table}[ht]
% \vskip -0.2in
\caption{Ablation on INR architectures. We report the $L_2$ relative error (\%) over the training horizon (\textsc{In-t}) and the subsequent duration (\textsc{Out-t}) in CE1 scenarios.}
\label{exp:ablation_inr}
% \vskip 0.15in
\begin{center}
\begin{small}
\begin{sc}
\resizebox{1.0\linewidth}{!}{
\begin{tabular}{lcccccccc}
\toprule
 \multirow{2}{*}{INR} & \multicolumn{2}{c}{PI-DeepONet} && \multicolumn{2}{c}{MAD} && \multicolumn{2}{c}{\name} \\ 
 \cmidrule{2-3} \cmidrule{5-6} \cmidrule{8-9}    
& In-t & Out-t && In-t & Out-t && In-t & Out-t  \\
\midrule
MLP (\textit{tanh}) & 15.25 & 27.72 && 4.24 & 17.06&& 2.66 & 6.40 \\ 
MLP (\textit{sin})  & 17.75 & 26.67 && 4.14 & 15.98&& 2.33 & 5.59 \\ 
FourierNet & \textbf{4.18} & \textbf{8.61} && \textbf{3.98} & \textbf{9.32} && \textbf{1.48} & \textbf{2.24} \\ 
\bottomrule
\end{tabular}}
\end{sc}
\end{small}
\end{center}
\vskip -0.2in
\end{table}

\subsection{Sample Efficiency}
\begin{figure}[ht]
% \vskip 0.2in
\begin{center}
\centerline{\includegraphics[width=0.7\linewidth]{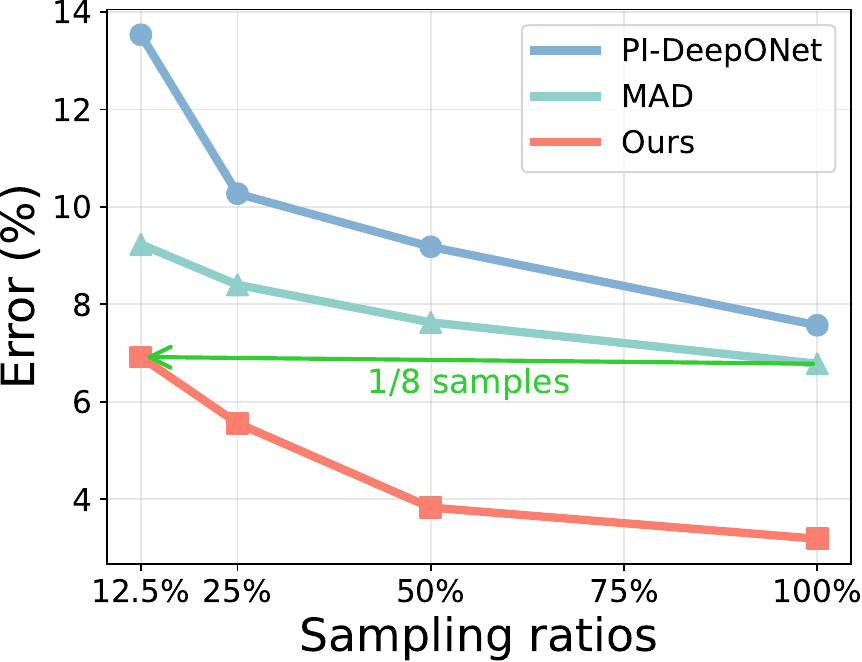}}
\vskip -0.1in
\caption{\textbf{CE3} test \textit{In-t} performance vs. numbers of
training pairs. }
\label{fig:training_pairs}
\end{center}
\vskip -0.2in
\end{figure}

We conduct a comparative analysis of {\name} against baseline models across varying numbers of training pairs. Specifically, we focus on the \textbf{CE3} scenario and sample subsets of training pairs with different ratios, denoted as $s\in\{12.5\%, 25\%, 50\%, 100\%\}$, where $s=100\%$ corresponds to the complete training set. We report results in the test set \textit{In-t} in~\cref{fig:training_pairs}. Our observations indicate that {\name} consistently achieves optimal performance across all sample ratios and exhibits reduced sensitivity to the reduction of training pairs in comparison to PI-DeepONet. 
Notably, {\name}, even when utilizing only 12.5\% of training pairs, achieves comparable performance with the other two baselines employing 100\% of the training pairs.

\subsection{Visualizations}
\begin{figure*}[ht]
% \vskip 0.2in
\begin{center}
\centerline{\includegraphics[width=0.7\textwidth]{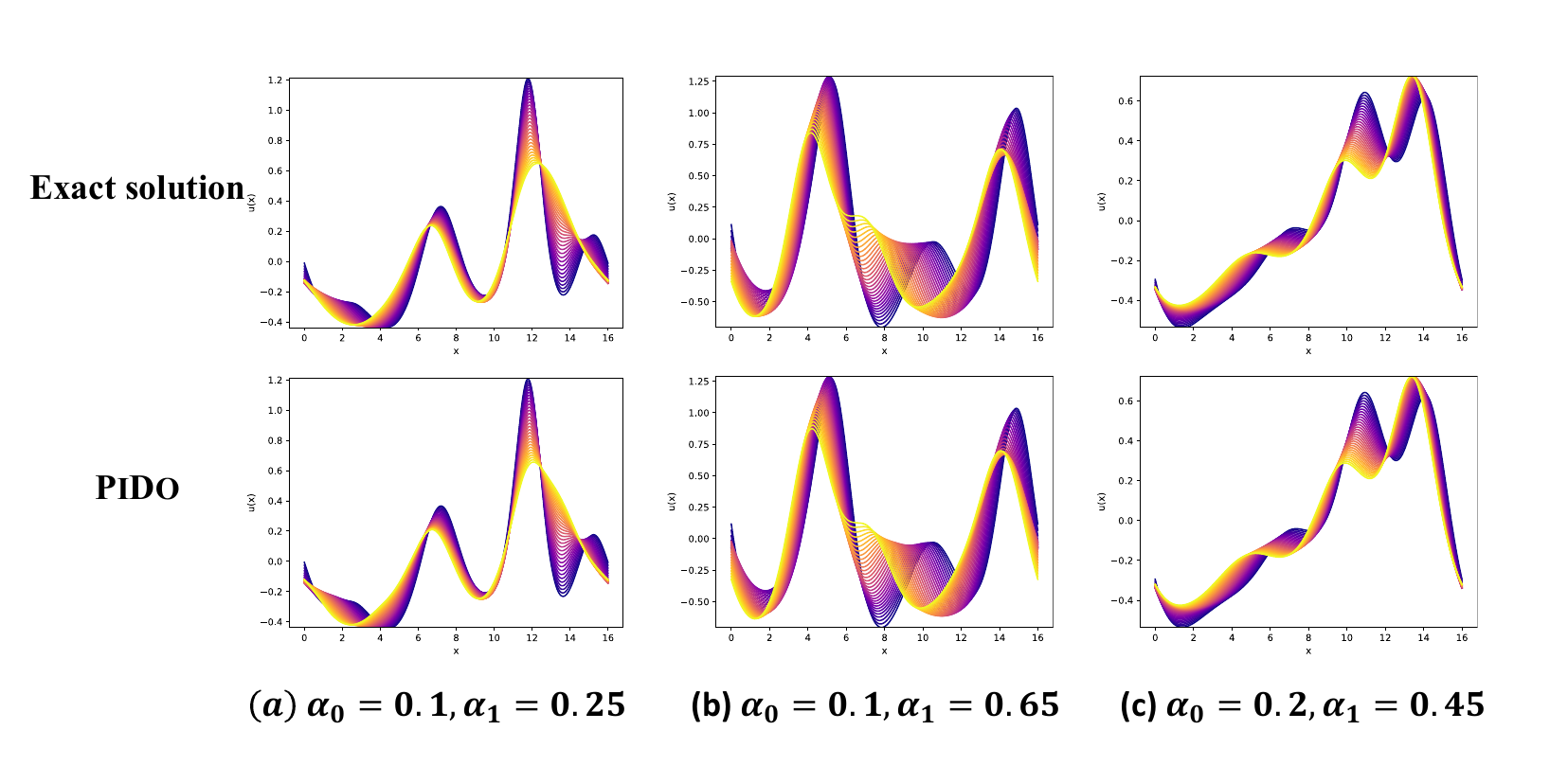}}
\vskip -0.2in
\caption{Prediction of {\name} on 1D combined equations with different $\boldsymbol{\alpha}$. }
\label{fig:ce3_prediction}
\end{center}
\vskip -0.2in
\end{figure*}

\begin{figure*}[ht]
% \vskip 0.2in
\begin{center}
\centerline{\includegraphics[width=0.7\textwidth]{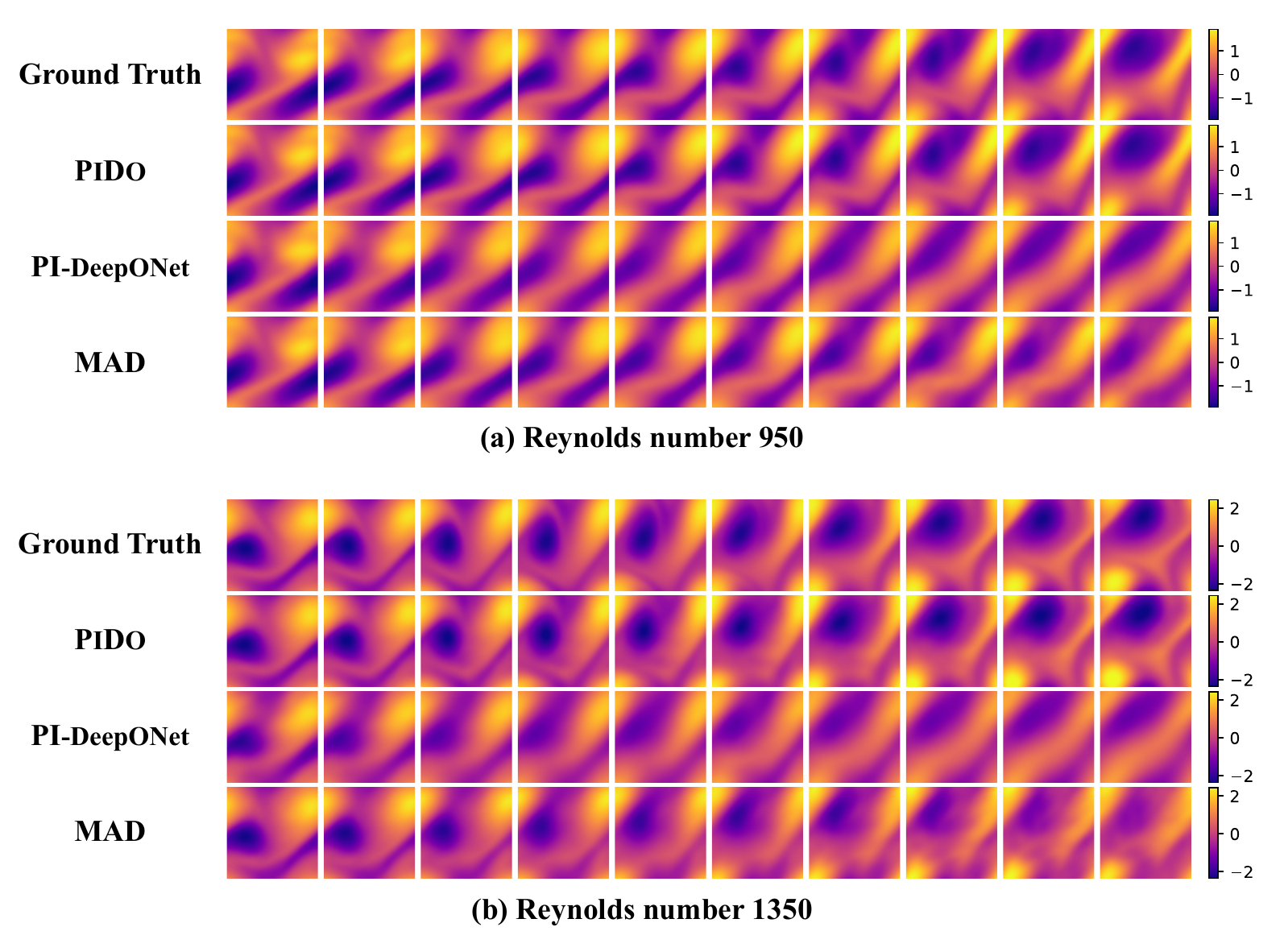}}
% \vskip -0.2in
\caption{Prediction of {\name} on 2D NS equations with different Reynolds number. The last 5 frames
are beyond the training horizon.}
\label{fig:ns2_prediction}
\end{center}
\vskip -0.2in
\end{figure*}

We visualize the predictions of {\name} for the \textbf{CE3} test set in~\cref{fig:ce3_prediction} and for the \textbf{NS2} test set in \cref{fig:ns2_prediction}.

\section{Auto-Decoder versus Auto-Encoder for Physics-Informed EDM}
\label{sec:detail_auto_encoder}
Since physics-informed training strategies for auto-encoders differ significantly from those used for auto-decoders (our method), this section discusses these key differences and details the specific training settings employed in auto-encoder methods.

Previous work PINODE~\cite{sholokhov2023physics} utilizes an auto-encoder framework for physics-informed training. Unlike our auto-decoder, which takes spatial coordinates $\boldsymbol{x}$ and embeddings $\boldsymbol{c}$ as input data and output predictions of states $\boldsymbol{u}$, PINODE's encoder $\mathcal{E}$ operates in the opposite direction. It takes $\boldsymbol{u}$ as input and output $\boldsymbol{c}=\mathcal{E}(\boldsymbol{u})$. This allows PINODE to calculate the derivative of $\boldsymbol{c}$ w.r.t $\boldsymbol{u}$ using Auto-Differentiation (AD). Consequently, PINODE can derive the temporal derivative of its embedding $\boldsymbol{c}$, assuming the input data $\boldsymbol{u}$ follows the PDE $\frac{\partial\boldsymbol{u}}{\partial t}=\mathcal{L}(\boldsymbol{u})$:
\begin{equation}
\label{eq:auto_encoder}
    \frac{\partial\boldsymbol{c}}{\partial t} = \frac{\partial\boldsymbol{c}}{\partial\boldsymbol{u}}\cdot\frac{\partial\boldsymbol{u}}{\partial t}
    =\frac{\partial\boldsymbol{c}}{\partial\boldsymbol{u}}\cdot(-\mathcal{L}(\boldsymbol{u})).
\end{equation} 
This derivative then serves as labels to train the dynamics model $\mathcal{F}$ by the following loss:
\begin{equation}
\label{eq:loss_auto_encoder}
    l_{\text{PDE}}(\theta_{\mathcal{E}}, \theta_{\mathcal{F}})=\|
    \frac{\partial\boldsymbol{c}}{\partial t}-\mathcal{F}(\boldsymbol{c})\|_2^2 = \|\frac{\partial\boldsymbol{c}}{\partial\boldsymbol{u}}\cdot(-\mathcal{L}(\boldsymbol{u}))-\mathcal{F}(\boldsymbol{c})\|_2^2,
\end{equation} 
where $\theta_{\mathcal{E}}$ and $\theta_{\mathcal{F}}$ denote parameters of $\mathcal{E}$ and $\mathcal{F}$, respectively.

However, this approach has three limitations. 
First, PINODE's encoder and decoder adheres to a fixed grid for both input data and output predictions, limiting its flexibility.
Second, PINODE relies on an analytical representation of the input data $\boldsymbol{u}$ to compute its spatial derivatives involved in $\mathcal{L}(\boldsymbol{u})$ in~\cref{eq:loss_auto_encoder}. 
This is achieved by pre-defining an analytical distribution from which the input data is sampled. 
Finally, PINODE assumes the input data distribution accurately reflects the true PDE solutions.
However, finding such a representative distribution in real-world scenarios can be challenging. 

Deviations from this assumption can significantly degrade PINODE's performance. Our experiments (\cref{exp:auto_encoder_ic}) demonstrate this sensitivity. 
In these experiments, we sample PINODE's input data from exact PDE solutions (ideal scenario) or the distribution of initial conditions (the only data we can access in the data-constrained scenario). The results show that PINODE's time extrapolation performance suffers significantly when the input data deviates from the true distribution of PDE solutions. 
Additionally, increasing the number of initial conditions offered little improvement, further highlighting PINODE's dependence on a suitable input data distribution. 
Given the data-constrained setting of our experiments in~\cref{sec:experiments}, we employ the distribution of initial conditions as input data for PINODE to ensure a fair comparison with other methods.

In contrast, {\name} adopts the auto-decoder framework, which is grid-independent and enables the calculation of spatial derivatives of predicted solutions through AD.
Moreover, the training of {\name} does not require any prior knowledge about data distribution, making it more robust for real-world scenarios.
\begin{table}[ht]
\caption{The performance of PINODE with different input data. $N_\text{IC}$ denotes the number of initial conditions. We report the $L_2$ relative error (\%) on the CE3 scenario.}

\label{exp:auto_encoder_ic}
% \vskip 0.15in
\begin{center}
\begin{small}
\begin{sc}
\resizebox{1.0\linewidth}{!}{
\begin{tabular}{llcc}
\toprule
 Method & Input data & In-t & Out-t\\ 
\midrule
PINODE & Exact solutions & 5.77& 11.02 \\
\midrule 
PINODE &  Initial conditions ($N_\text{IC}$=3584) & 18.21& 39.41 \\
PINODE & Initial conditions ($N_\text{IC}$=7168) & 18.25& 41.26 \\
PINODE & Initial conditions ($N_\text{IC}$=14336)& 17.74& 39.19 \\
{\name}& Initial conditions ($N_\text{IC}$=3584) & 3.19 & 8.08 \\
\bottomrule
\end{tabular}}
\end{sc}
\end{small}
\end{center}
\vskip -0.2in
\end{table}

\section{Experiments Setups}
\label{sec:exp_setup}
\subsection{Training Details}
\label{sec:training_details}
\paragraph{1D combined equations} For this problem, we consider the periodic boundary condition and construct training and test sets with initial conditions sampling from the super-position of sinusoidal waves given by $\sum_{i=1}^N A_i \mathrm{sin}(\frac{2\pi k_i}{L_x}x+b_i)$, where $\{A_i\}$, $\{b_i\}$ and $\{k_i\}$ denote random amplitudes, phases and integer wave numbers. We set $L_x=16$, $T_{tr}=0.96s$ and $T=1.92s$. 
We employ a uniform spatial discretization of 400 cells encompassing the interval [0, 16). The temporal domain is discretized into 60 time steps using a uniform spacing over the interval [0, 1.92]. During training, we sample collocation points from this grid for physics-informed training.
{\name} is trained for 3000 epochs with a batch size of 128 and a learning rate of 1e-3. The weights of alignment and smoothing regularization are set to 1 and 0.01, respectively.
We consider three scenarios for this equation:
\begin{itemize}
    \item \textbf{CE1} for Burgers’ equation with $\boldsymbol{\alpha}=(0.1,0)$, generating initial conditions with $A_i\in[-0.5, 0.5]$, $k_i\in\{1, 2\}$, $b_i\in[0, 2\pi]$ and N=2; We generate 3584 trajectories for training and 512 trajectories for testing. 
    \item \textbf{CE2} for KdV equation with $\boldsymbol{\alpha}=(0,0.05)$, generating initial conditions with $A_i\in[-0.5, 0.5]$, $k_i\in\{1, 2\}$, $b_i\in[0, 2\pi]$ and N=2; We generate 3584 trajectories for training and 512 trajectories for testing.  
    \item \textbf{CE3} for combined equation with $\boldsymbol{\alpha}\in\{(\alpha_0, \alpha_1)|0<\alpha_0\leq0.4,0<\alpha_1\leq0.65\}$. Specifically, we use the training set $\boldsymbol{\alpha}_{tr}\in\{0.1,0.2,0.3,0.4\}\times\{0.05,0.25,0.45,0.65\}$ and the test set  $\boldsymbol{\alpha}_{ts}\in\{0.15,0.25,0.35\}\times\{0.15,0.35,0.55\}$. We generate initial conditions with $A_i\in[0, 1]$, $k_i\in\{1, 2\}$, $b_i\in[0, 2\pi]$ and N=2; We generate 224 trajectories for each configuration of training coefficients (3584 in total) and 32 trajectories for each configuration of testing coefficients (288 in total).
\end{itemize}

\paragraph{2D Navier-Stokes equation} 
For this problem, trajectories are simulated under periodic boundary conditions, employing initial conditions described in \cite{li2020fourier}. 
We set $T_{tr}=5s$ and $T=10s$. We employ a uniform spatial discretization of 64*64 cells encompassing the interval $[0, 1)^2$. 
We consider a temporal domain of [0, 10] and use a time step size of 0.5 seconds for PI-DeepONet and MAD to ensure training stability. For {\name}, we use a time step size of 1 second, benefiting from the latent dynamics smoothing.
We sample collocation points from this grid for physics-informed training. 
For {\name}, the weights of alignment and smoothing regularization are set to 1 and 0.01, respectively.
We consider two scenarios:
\begin{itemize}
    \item \textbf{NS1} for fixed Reynolds number $\alpha=1000$. We generate 1024 trajectories for training and 128 trajectories for testing. {\name} is trained for 3000 epochs with a batch size of 16 and a learning rate of 2e-3. For the dynamics model of {\name}, the learning rate is set to 2e-4.   
   \item \textbf{NS2} for diverse Reynolds numbers. We utilizes a training set encompassing $\alpha$ values sampled from $\{700, 800, 900, 1000, 1100, 1200, 1300, 1400\}$ and a testing set incorporating Reynolds numbers from $\{750, 850, 950, 1050, 1150, 1250, 1350\}$. We generate 256 trajectories for each configuration of training coefficients and 32 trajectories for each configuration of testing coefficients. {\name} is trained for 6000 epochs with a batch size of 64 and a learning rate of 1e-3. For the dynamics model of {\name}, the learning rate is set to 1e-4.   
\end{itemize}

\subsection{Implementations}
\label{sec:implment}
\paragraph{{\name}} The decoder is a FourierNet~\cite{fathony2020multiplicative} with 3 hidden layers and a width of 64. We opted to employ FourierNet due to its demonstrated superior performance in tasks similar to ours. To maintain a fair and controlled comparison with baseline methods, we utilize the same network architecture across all approaches in this work. 
An FourierNet with k hidden layers is defined via the following recursion
\begin{equation*}
    \begin{split}
        \boldsymbol{z}^{(1)} =& \text{sin}(\omega^{(1)}\boldsymbol{x}+\phi^{(1)}),
        \,\\
        \boldsymbol{z}^{(i+1)} =& (W^{(i)}\boldsymbol{z}^{(i)}+b^{(i)})\\&\circ\, \text{sin}(\omega^{(i+1)}\boldsymbol{x}+\phi^{(i+1)}),\,i=1,2,...,k,\\
        \boldsymbol{z}_{\text{out}}=&W^{(k+1)}\boldsymbol{z}^{(k+1)}+b^{(k+1)},
    \end{split}
\end{equation*}
where $\boldsymbol{x}$ is the input coordinates, $\circ$ is the elemental multiplication and $\{W^{(i)},b^{(i)},\omega^{(i)},\phi^{(i)} \}$ denote the trainable parameters. To incorporate embedding $\boldsymbol{c}$ into a FourierNet, we modulate both the amplitudes and phases of the sinusoidal waves generated by the hidden layers:
\begin{equation*}
    \begin{split}
        \boldsymbol{z}^{(i+1)} = & (W^{(i)}\boldsymbol{z}^{(i)}+b^{(i)} + W_{A}^{(i)}\boldsymbol{c})\\
        &\circ\, \text{sin}(\omega^{(i+1)}\boldsymbol{x}+\phi^{(i+1)}+W_{P}^{(i)}\boldsymbol{c})),\,i=1,2,...,k.
    \end{split}
\end{equation*}
The dynamics model is a 4-layer MLP with a width of 512. The activation function of dynamics model is Swish. We use the RK4 integrator via TorchDiffEq~\cite{torchdiffeq} for the training of dynamics model. We set the code size to 64 for 1D combined equations and to 128 for 2D NS equations.

\paragraph{PI-DeepONet} The trunk net is a FourierNet with 3 hidden layers and a width of 64. The branch net is a 4-layer SIREN~\cite{sitzmann2020implicit} with a width of 512. We tune the learning rate within the range \([1\text{e-4}, 5\text{e-3}]\) and report the best results.

\paragraph{PINODE} Both the encoder and decoder are SIRENs with 3 hidden layers and a width of 64. The encoder takes the discrete initial conditions as input, which are vectors of 400 elements for 1D problems and 4096 elements (64*64) for 2D problems, and generates the latent embedding. The decoder operates in the opposite direction. The dynamics model and latent code configurations are identical to those employed in our {\name} method.

\paragraph{MAD} The decoder is a FourierNet with 3 hidden layers and a width of 64. We set the embedding size to 64 for 1D combined equations and to 128 for 2D NS equations. When inference on new initial conditions or PDE coefficients, we only finetune the learnable embeddings while freeze the parameters of decoder. We tune the learning rate within the range \([1\text{e-4}, 5\text{e-3}]\) and report the best results.

\subsection{Computational Efficiency}
\label{sec:com_eff}
We used 4 NVIDIA RTX3090 GPUs for all experiments. We compare the computational time and memory usage of {\name} and the baselines in the CE1 setting in~\cref{exp:compute}.
While {\name} incurs slightly higher memory and computation demands due to the use of Neural ODE, this trade-off is demonstrably worthwhile. The resulting performance boost is significant, and overall resource requirements remain relatively low.

\begin{table}
% \vskip -0.15in
\caption{The computational time and memory usage of each method.}
% \vskip -0.15in
\label{exp:compute}
% \vskip 0.15in
\begin{center}
\begin{small}
\begin{sc}
% \resizebox{1.0\linewidth}{!}{
\begin{tabular}{c|cc}
\toprule
Method & Sec/epoch & MB/sample \\ 
\midrule
PI-DeepONet & 27 & 533 \\
MAD & 29 & 569 \\
{\name} & 35 & 582 \\
\bottomrule
\end{tabular}
\end{sc}
\end{small}
\end{center}
\vskip -0.3in
\end{table}

% {\appendix[Proof of the Zonklar Equations]
% Use $\backslash${\tt{appendix}} if you have a single appendix:
% Do not use $\backslash${\tt{section}} anymore after $\backslash${\tt{appendix}}, only $\backslash${\tt{section*}}.
% If you have multiple appendixes use $\backslash${\tt{appendices}} then use $\backslash${\tt{section}} to start each appendix.
% You must declare a $\backslash${\tt{section}} before using any $\backslash${\tt{subsection}} or using $\backslash${\tt{label}} ($\backslash${\tt{appendices}} by itself
%  starts a section numbered zero.)}

%{\appendices
%\section*{Proof of the First Zonklar Equation}
%Appendix one text goes here.
% You can choose not to have a title for an appendix if you want by leaving the argument blank
%\section*{Proof of the Second Zonklar Equation}
%Appendix two text goes here.}

\bibliography{IEEEabrv,./reference}
\bibliographystyle{IEEEtran}

\vfill

\end{document}